\documentclass[10pt,twocolumn,letterpaper]{article}

\usepackage{cvpr}              

\definecolor{cvprblue}{rgb}{0.21,0.49,0.74}
\usepackage[pagebackref,breaklinks,colorlinks,allcolors=cvprblue]{hyperref}
\usepackage{multirow}
\usepackage[normalem]{ulem}
\useunder{\uline}{\ul}{}
\usepackage{colortbl}
\usepackage{float}
\restylefloat{table}
\usepackage{makecell}
\usepackage{tabularx}

\newcommand{\method}{AnomalyAny}

\def\paperID{14868} 
\def\confName{CVPR}
\def\confYear{2025}

\title{
Unseen Visual Anomaly Generation
}

\author{Han Sun\\
EPFL, Switzerland\\
\and
Yunkang Cao\\
HUST, China\\
\and
Hao Dong\\
ETH Zurich, Switzerland\\
\and
Olga Fink\\
EPFL, Switzerland\\
\and
\vspace{-1cm}
\tt\small \{han.sun, olga.fink\}@epfl.ch, cyk\_hust@hust.edu.cn, hao.dong@ibk.baug.ethz.ch
}

\begin{document}
\maketitle
\begin{abstract}

Visual anomaly detection (AD) presents significant challenges due to the scarcity of anomalous data samples. 
While numerous works have been proposed to synthesize anomalous samples, these synthetic anomalies often lack authenticity or require extensive training data, limiting their applicability in real-world scenarios.
In this work, we propose \textbf{Anomaly Anything} (\method{}), a novel framework that leverages Stable Diffusion (SD)'s image generation capabilities to generate diverse and realistic unseen anomalies. 
By conditioning on a single normal sample during test time, \method{} is able to generate unseen anomalies for arbitrary object types with text descriptions.
Within \method, we propose attention-guided anomaly optimization to direct SD’s attention on generating hard anomaly concepts. Additionally, we introduce prompt-guided anomaly refinement, incorporating detailed descriptions to further improve the generation quality. 
Extensive experiments on MVTec AD and VisA datasets demonstrate \method's ability in generating high-quality unseen anomalies and its effectiveness in enhancing downstream AD performance.
Our demo and code are available at \href{https://hansunhayden.github.io/AnomalyAny.github.io/}{https://hansunhayden.github.io/AnomalyAny.github.io/}.
\end{abstract}    
\section{Introduction}
\label{sec:intro}



\begin{figure*}
    \centering
    \includegraphics[width=0.95\textwidth]{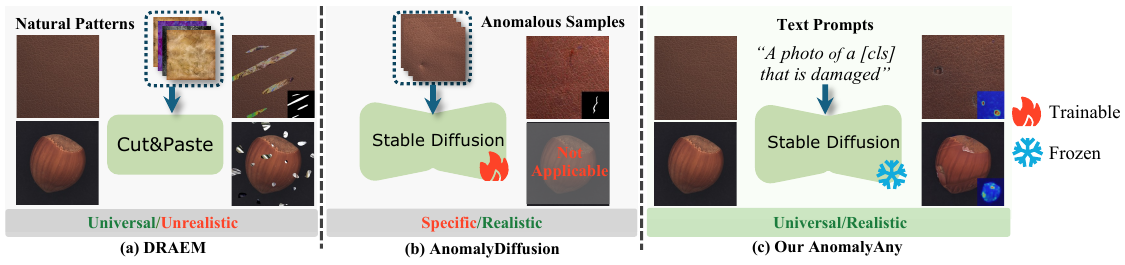}
   \vspace{-0.4cm}
    \caption{\textbf{Comparison between different visual anomaly generation methods.} In comparison to existing methods, \method{} can generate diverse and realistic unseen anomalies without training.}
    \vspace{-0.2cm}
    \label{fig:intro}
\end{figure*}

Visual anomaly detection (AD)~\cite{cao2024survey} aims to identify unusual or unexpected patterns in image data that deviate from a normal distribution of features, and is essential in fields such as industrial inspection and quality control~\cite{MVTec-AD}. 
Since anomalies are rare and difficult to collect, most existing AD methods rely on unsupervised learning ~\cite{uniad,HVQ-Trans,fastrecon} using solely normal samples.
Despite recent advancements in the AD field, the scarcity of anomalous samples for training remains a persistent challenge.
To address this, various studies have explored visual anomaly generation, as illustrated in~\cref{fig:intro}. 
Some approaches \cite{zavrtanik2021draem, zhang2024realnet} augment normal samples by cropping and pasting random patterns, either from natural patterns in other datasets or from the images themselves. While this generates diverse anomalous samples and is applicable to unseen objects without training, these samples often appear unrealistic.
Another approach involves generating visual anomalies using generative models \cite{hu2024anomalydiffusion}, which produce more realistic images but require sufficient and representative normal and abnormal samples for training, which is typically challenging to obtain for AD. Given the rarity and variability of anomalies, collecting a representative set of anomalous samples is difficult. Additionally, the diversity of product variants and configurations in industrial applications often results in a lack of representative normal samples~\cite{WinClip}, presenting an additional challenge. These limitations lead to a shortage of both abnormal and normal samples.
Consequently, generative models are often not suitable for real-world applications with limited data and may be biased towards the few available training observations.

Given the limitations of existing visual anomaly generation methods, we aim to generate diverse and realistic unseen anomalies using minimal normal data \textit{without} any anomalous samples. The goal motivates us to explore Stable Diffusion (SD) \cite{rombach2022high}, a
latent text-to-image diffusion model known for generating diverse images across various domains.
Although SD demonstrates impressive image generation capabilities, it is not specifically designed for visual anomaly generation. Consequently, when applied directly for anomaly generation, SD may produce images that deviate from the desired normal distribution and fail to accurately represent realistic anomaly patterns (\cref{fig:attention} (b)). 
Prior approaches~\cite{zhang2024realnet,hu2024anomalydiffusion} suggest fine-tuning SD on available normal or anomalous samples, but this approach is constrained in data scarcity scenarios and can compromise SD's ability to generalize to unseen data and anomaly types.

In this work, we propose \textbf{Anomaly Anything} (\method{}) for realistic and diverse unseen visual anomaly generation.
We leverage SD for unseen visual anomaly generation with anomaly text descriptions, which can be defined manually or automatically acquired from GPT-4~\cite{openai2024gpt4} with object category as input.
Instead of fine-tuning SD on normal data, we introduce \textit{\textbf{test-time normal sample conditioning}}, guiding the generation process on a single normal sample. This approach preserves the diversity and generalization ability of SD, enabling anomaly generation on novel data and new anomaly types. 
With a single model, we can generate realistic and diverse anomalous samples for any new objects and from arbitrary anomaly descriptions. 
We observe that the original SD struggles to generate realistic anomaly samples due to two inherent challenges: first, anomalies are comparatively rare in SD's training data; second, unlike common objects and patterns, anomalous patterns typically occupy only a small region of the image, making them easily overlooked in generation. 
To address this, we propose \textit{\textbf{attention-guided anomaly optimization}} to enforce the model's attention on generating anomaly concepts by maximizing the attention value correlated with the anomaly token.
To further guide and refine the generation results, we propose \textit{\textbf{prompt-guided anomaly refinement}}, which leverages more detailed anomaly descriptions as additional semantic guidance. 
Our main contributions are:

\begin{itemize}

    \item We propose \method{}, a visual anomaly generation framework for unseen anomalies.
    Users can provide arbitrary normal images of objects and anomaly descriptions to generate realistic and diverse anomalous samples.

    \item  We introduce attention-guided anomaly optimization along with prompt-guided anomaly refinement to overcome the limitations of SD in anomaly generation. Our proposed method achieves more authentic generation results than existing methods, without requiring additional training or abundant normal/abnormal data samples.
    \item We demonstrate the effectiveness of our proposed \method{} in both generation quality and in facilitating the training of the downstream anomaly detection models.
\end{itemize}

\section{Related Works}

\subsection{Anomaly Generation}
The scarcity of anomalous data has motivated numerous studies to synthesize anomaly samples~\cite{nngmix}.
One approach augments normal training samples by cropping and pasting abnormal patterns from few-shot abnormal samples in the test set~\cite{snell2017prototypical, lin2021few}
, natural patterns of external datasets~\cite{cimpoi2014describing, zavrtanik2021draem}, or directly from the normal image itself~\cite{li2021cutpaste, schluter2022natural}. Despite being simple and effective, these methods lack authenticity and diversity in the generated samples.
Another approach employs generative models, such as generative adversarial networks (GANs) to synthesize anomalies~\cite{zhang2021defect, duan2023few}. More recently, advancements in diffusion models~\cite{hu2024anomalydiffusion, yang2025defect,AnoGen,zhang2024realnet} have enabled methods that fine-tune these models on anomaly data to produce more diverse and realistic anomalies. 
However, these methods typically require substantial amounts of normal and/or abnormal data samples to learn the dataset-specific distribution, which is often impractical in data-limited scenarios. Also, these methods can only generate samples similar to the training set, \ie, seen anomalies, but fail in generating unseen anomalies. In real-world applications, however, it is often impractical to assume prior knowledge of all possible anomalies, highlighting the need for methods that can generalize to unseen types.
To generate unseen yet realistic anomalies, \method{} directly deploys the pretrained SD model for anomaly synthesis without additional training,  offering a more flexible solution for high-quality anomaly generation across arbitrary objects.


\subsection{Anomaly Detection}

The scarcity of available anomalous samples has made \textbf{unsupervised AD}~\cite{RD4AD,CDO,Patchcore} the dominant paradigm in the AD field, where the goal is to model the normal distribution and then identify anomalies as outliers~\cite{RD4AD,CDO,Patchcore}.
However, these methods rely on the availability of a sufficient amount of normal samples that adequately represent the underlying distribution. Due to the lack of representative normal samples across the wide range of product variations and industrial configurations, \textbf{few-shot AD} has gained increasing attention~\cite{uniad,HVQ-Trans}. 
Existing few-shot methods typically leverage additional knowledge from vision-language models pretrained on large-scale datasets, such as Contrastive Language-Image Pre-training (CLIP)~\cite{clip}, to compute similarities between data samples and the normal or abnormal text prompts~\cite{WinClip, anomalygpt, promptad}. 
Nevertheless, training AD models solely with normal samples remains challenging due to a lack of awareness of anomaly distributions, underscoring the need for realistic anomalous samples to enhance AD performance.

\begin{figure*}[t]
    \centering
    \includegraphics[width=1.0\textwidth]{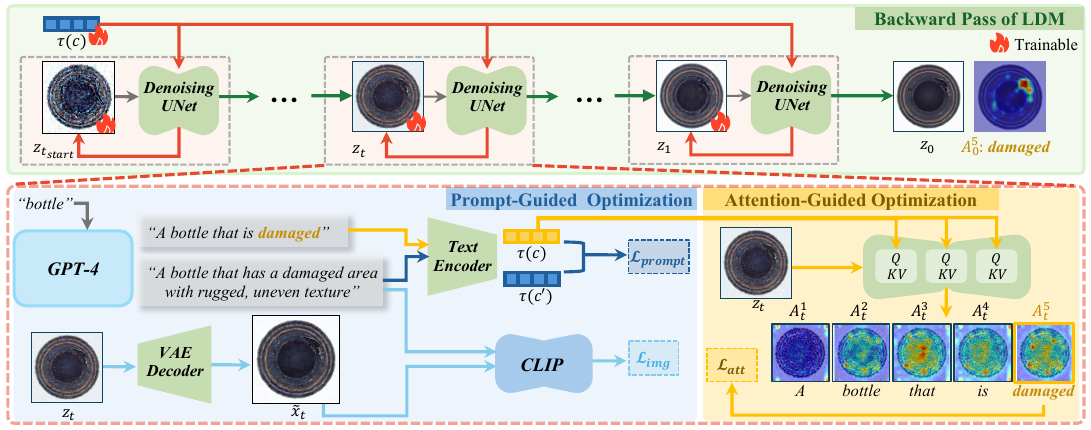}
    \vspace{-0.8cm}
    \caption{\textbf{Illustration of \method{}} with details of the attention-guided \& prompt-guided optimization process at time step $t$. }
    \label{fig:att_optimization}
    \vspace{-0.2cm}
\end{figure*}

\section{Preliminary: Stable Diffusion}\label{DDPM}
\textbf{Diffusion models.} Denoising diffusion probabilistic models (DDPMs) \cite{ho2020denoising} learn the desired data distribution by defining a Markov chain of length $T$.
In the forward pass, this chain gradually adds noise to a given data sample $x_0$ to obtain a sequence of noisy samples $x_t, t\in T$.
In the reverse process, a model $\epsilon_\theta$ parameterized by $\theta$ is learned to predict the noise added for each step $t$.
Our method is based on SD~\cite{rombach2022high}, a latent diffusion model (LDM) that applies the denoising diffusion process to the latent representation $z$ of $x$ in the latent space of a variational auto-encoder (VAE) with the learning objective of predicting the added noise at each time step t as:
\begin{equation}
L_{L D M}:=\mathbb{E}_{\mathcal{E}(x), \epsilon \sim \mathcal{N}(0,1), t}\left[\left\|\epsilon-\epsilon_\theta\left(z_t, t\right)\right\|_2^2\right],
\end{equation}
where $z_t$ represents the noised latent representation at time step $t$.
During inference, the reverse process starts with a random noise $x_T \sim \mathcal{N}(0, \mathbf{I})$ and gradually generates an image sample from the noise from step $T$ to $0$.


\noindent\textbf{Text condition.} SD introduces text guidance via a cross-attention mechanism. The denoising UNet network in the latent space of SD consists of self-attention layers followed by cross-attention layers at resolutions $ P \in (64, 32, 16, 8)$. 
Given a text prompt $c$ composed of $N$ tokens, a guidance vector $\tau(c)$ is then obtained via the CLIP text-encoder $\tau$   \cite{radford2021learning}.
$\tau(c)$ is then mapped to intermediate feature maps of the DDPM model $\mathbf{\epsilon_\theta}$ through  each cross-attention layer as follows:
\begin{small}
\begin{align}\label{att}
    A &=\operatorname{softmax}\left(\frac{Q K^T}{\sqrt{d}}\right) \cdot V, 
    \quad A \in [P, P, N] ,\\
    Q&=W_Q^{(i)} \cdot \varphi_i\left(z_t\right), K=W_K^{(i)} \cdot \tau(c), V=W_V^{(i)} \cdot \tau(c) ,
\end{align}
\end{small}

\noindent where $\varphi_i(z_t)$ is the intermediate feature of the UNet. $A, Q, K, V$ denote the attention, query, key, and value matrices. $W^{(i)}$ represents learnable projection matrices. 
Based on image-captioning pairs $\{x, c\}$, the text-conditioned diffusion model can be learned with the objective:
\begin{small}
\begin{equation}\label{ldm}
L_{L D M}:=\mathbb{E}_{\mathcal{E}(x), c, \epsilon \sim \mathcal{N}(0,1), t}\left[\left\|\epsilon-\epsilon_\theta\left(z_t, t, \tau(c)\right)\right\|_2^2\right] .
\end{equation}
\end{small}

\begin{figure*}[t]
  \centering
  \includegraphics[width=0.8\linewidth]{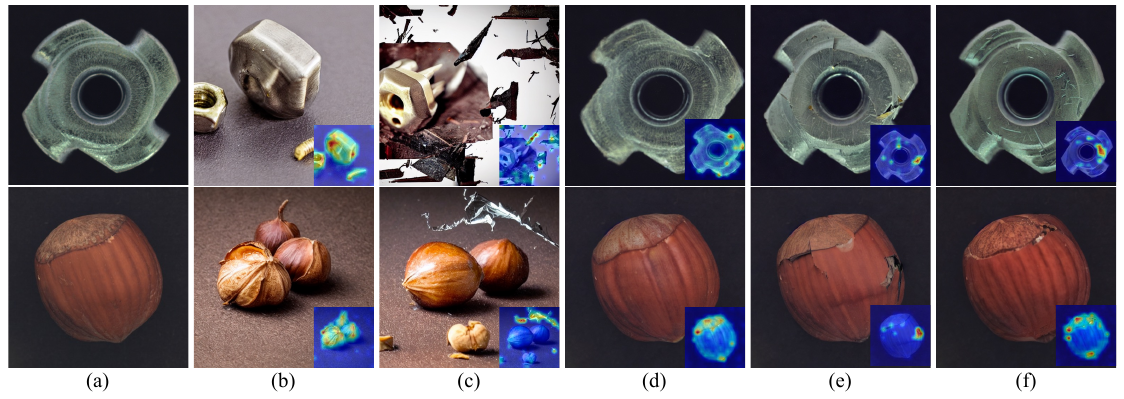}
  \vspace{-0.2cm}\caption{\textbf{Examples of generated anomaly samples and attention map of \textit{damage}.} We present (a) Normal guidance image, and the results generated by (b) Stable Diffusion, (c) Ours w/o normal sample conditioning, (d) Ours w/o attention-guided optimization, 
    (e) Ours w/o prompt-guided optimization, and (f) Our proposed \method.}
    \vspace{-0.4cm}
    \label{fig:attention}
\end{figure*}

\section{Methodology}

In this work, we propose \method{}, a novel framework for generating realistic, promptable anomalies on unseen objects and anomaly types, as illustrated in \cref{fig:att_optimization}. \method{} comprises three core modules. The framework begins with Test-time Normal Sample Conditioning (\cref{normal_condition}) to guide the generation process on a single normal sample, enabling anomaly generation on novel objects and anomaly types. During generation, we introduce a two-stage optimization process: Attention-Guided Anomaly Optimization (\cref{atten}) first focuses generation on hard anomaly concepts, followed by Prompt-guided Anomaly Refinement (\cref{refine}) that leverages detailed text descriptions to further enhance the generation quality. 

\subsection{Test-time Normal Sample Conditioning}\label{normal_condition}

In our work, we leverage SD to generate unseen visual anomalies. 
While SD is effective for image generation, the inherent diversity of its underlying distribution causes the generated images to significantly differ from the normal images in a specific AD dataset, as shown in \cref{fig:attention} (b).
To generate images that align with the target normal distribution, instead of fine-tuning SD, we incorporate information from normal samples directly during the generation process. Given a normal sample $x^{normal}$ and its latent representation $z^{normal} = \mathcal{E}(x^{normal})$ encoded by the VAE encoder, we get a sequence of samples $z_1^{normal}, z_2^{normal}, ..., z_T^{normal}$ with progressively added noise controlled by the DDPM noise scheduler.
Instead of starting the inference from $z_T \sim \mathcal{N}(\mathbf{0}, \mathbf{I})$, we begin from step $t_{start} = T\cdot (1-\gamma)$ with the noisy latent representation $z_{t_{start}}^{normal}$,
which conveys corrupted features from the guidance normal image $x^{normal}$. This enables the generation of anomalies for novel object types during test time without requiring additional training.
The parameter $\gamma$ controls the starting step, i.e., the noise scale added to the guidance data sample.  
In our experiments, we set $\gamma = 0.25$ to balance similarity to the original distribution with inference steps that promote diverse outcomes. 

For more precise control conditioned on a given normal sample,  we use a mask input as an optional constraint to specify the location of the generated anomalies.  This mask can function either as a foreground mask to restrict anomalies to the intended object or as a detailed, pre-defined mask that accurately defines the anomaly region. For the predicted latent representation $z_t$ at time step $t$, we apply the following operation:
\begin{equation}
    z_t = mask \odot z_t + (1-mask)\odot z_t^{normal} .
\end{equation}
This formulation preserves the distribution of the normal guidance sample outside the masked region, ensuring that modifications are restricted solely to the specified area.



\subsection{Attention-Guided Anomaly Optimization}
\label{atten}

\begin{figure}
\centering
  \includegraphics[width=1.0\linewidth]{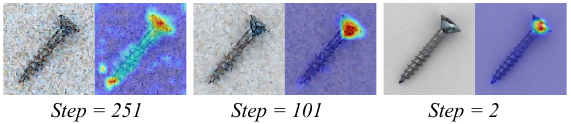}
  \vspace{-0.6cm}
    \caption{Visualization of \textbf{intermediate generation results} 
    and \textbf{the attention maps} of anomaly tokens 
    at different denoising steps.}
    \vspace{-0.3cm}
    \label{fig:optimization}
\end{figure}

We generate anomalous samples of a specific object type, denoted as $[cls]$ (e.g., bottle), by providing anomaly text descriptions, such as \textit{"A photo of a $[cls]$ that is damaged".} 
Due to the challenging nature of anomaly generation as introduced, the desired anomaly semantics are often neglected in the generated images, as shown in \cref{fig:attention} (d). 
To overcome this challenge, we introduce \textit{\textbf{attention-guided anomaly optimization}} to enforce the generation of these critical yet difficult-to-capture anomaly concepts. 
First, we aggregate the attention maps from SD for subsequent optimization.
As introduced in Sec.~\ref{DDPM}, SD introduces text guidance using a cross-attention mechanism. Given a text prompt $c$ composed of $N$ tokens, at each inference step $t$, we obtain a collection of attention maps $A_t \in P \times P \times N$ corresponding to $c$ at different resolutions. These attention maps capture the cross-correlation with each text token, where $A[:,:,i]$ represents the probability assigned to token $c_i, i\in (1, N)$.
Following the methodology of~\cite{chefer2023attend}, we average all the attention maps at a resolution of $16 \times 16$, as these are identified to be the most semantically informative. The resulting attention map $\Bar{A}_t$ is then normalized across the token dimension  and smoothed using  a $\text{Gaussian}(\cdot)$ function:
\begin{equation}
    \Bar{A}_t = \text{Gaussian}(\operatorname{softmax}(\Bar{A}_t)), \quad \Bar{A}_t \in 16\times 16 \times N.
\end{equation}
We enforce the generated image to convey the semantic meaning of the anomaly type token $c_j$ from the given $c$, e.g., \textit{"damaged"} in \textit{"A photo of a $[cls]$ that is damaged"} via optimization guided by $\Bar{A}_t$.
During the generation process, at each time step $t$, we optimize the intermediate latent representation $z_t$ by maximizing the attention associated with the anomaly description. The optimized latent representation is then denoised to obtain $z_{t-1}$, which is used in the subsequent optimization and denoising steps.
{as shown in \cref{fig:att_optimization}}.
Specifically, during optimization at step $t$, we extract the attention map $\Bar{A}_t^j = \Bar{A}_t[:, :, j]$ correlated with the anomaly token $c_j$. We then compute the gradient update for $z_t$ by applying the loss function $\mathcal{L}_{att}$ to $\Bar{A}_t^j$.
Furthermore, we increase controllability by allowing masks as optional inputs and restricting the optimization of $z_t$ to within the specified mask.  
The optimization at step $t$ is formulated as follows:
\begin{align}
\mathcal{L}_{att}=1-\max \left(\Bar{A}_t^j\odot \text{mask}\right), \\
\quad
 z_t \leftarrow z_t-\alpha_t \cdot \nabla_{z_t} \mathcal{L}_{att}\odot \text{mask} ,
\end{align}
where $\alpha_t$ is a scalar defining the step size. This objective encourages a high maximum value of $\Bar{A}_t^j$ to strengthen the activations of the anomaly token. 
Through iterative optimization, we progressively integrate the semantic characteristics of anomalies into the generated image, as illustrated in \cref{fig:optimization}.

\begin{figure}
  \includegraphics[width=1.0\linewidth]{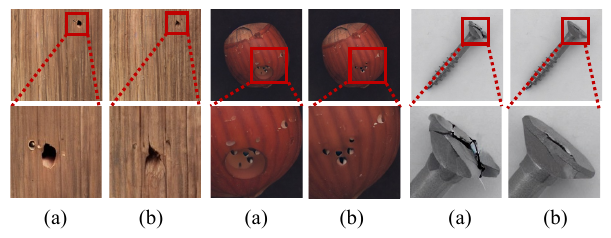}
    \vspace{-0.7cm}\caption{Examples of \textbf{anomalies} generated after attention-based optimization on anomaly tokens (a) w/o and (b) w/ localization-aware scheduler.}
    \vspace{-0.4cm}
    \label{fig:overfit}
\end{figure}

Our empirical results demonstrate that the iterative updates described above can cause redundant attention on specific pixels, resulting in image artifacts.
To mitigate over-optimization, we propose a \textit{\textbf{localization-aware scheduler}}. Starting with the initial generation result $z_{t_{start}}$ and its anomaly attention map $\Bar{A}_{t_{start}}^j$, we determine the number of activated pixels $n_{t_{start}}$ by counting pixels with attention values exceeding the mean in the smoothed attention map. At each optimization step $t$, we calculate  the activated pixel count $n_t$ and compute the scalar $\alpha_t$ as follows:
\begin{equation}\label{eq:optimization}
    \alpha_t = \lambda(1 + \Delta_t \cdot t)\cdot \frac{n_t}{n_{t_{start}}} ,
\end{equation}
where $\lambda$ is a scaling factor that controls the strength of the optimization, and $\Delta t$ adjusts the step size to gradually decrease the update rate. 
As the activated pixels become increasingly localized, we reduce  $\alpha_t$ to mitigate overfitting.
As illustrated in \cref{fig:overfit}, this strategy significantly mitigates unrealistic artifacts in the generated samples.




\subsection{Refined Anomaly Generation via Prompt-Guided Optimization}
\label{refine}

Attention-based optimization enforces the generation of anomaly keywords consisting of one to two tokens. However, the limited token length often results in ambiguous descriptions, thereby reducing the authenticity and richness of the generated context. To enhance semantic guidance and improve the generation quality, we propose a \textit{\textbf{prompt-guided optimization for anomaly refinement}} that incorporates detailed anomaly descriptions.
To enrich the generated anomaly distribution, we leverage GPT-4~\cite{achiam2023gpt} to generate potential anomaly types $c_j$ for a given object, along with their corresponding detailed descriptions $c'$. For example, given the object as \textit{bottle}, we prompt  GPT-4 to identify possible anomaly types and their descriptions, such as \textit{"damaged" $\rightarrow$ "close-up of a bottle that has a damaged area with rugged, uneven texture"}, thereby providing more nuanced guidance for generating a diverse range of damage types or anomalies.

To incorporate the semantic guidance from these long-text descriptions during the generation process, we introduce a CLIP-based image generation loss in the final denoising steps to ensure semantic consistency between the generated image and text guidance. At time step $t$, given the latent representation $z_t$, we obtain the decoded generation result $\tilde{x}_t = \mathcal{D}(z_t)$. We then minimize its distance from $c'$ in the CLIP~\cite{radford2021learning} embedding space as follows: 
\begin{align}
\mathcal{L}_{img}=1.0 - \operatorname{cosine}(\Phi^T(c'),  \Phi^V(\tilde{x}_t)) ,
\end{align}
where $\Phi^T(\cdot)$ and $\Phi^V(\cdot)$ denotes the CLIP text and visual encoders, respectively. Minimizing this similarity loss aligns the generated anomalies more closely with the semantic attributes of $c'$, refining the anomaly type with detailed descriptions. We then jointly optimize $z_t$  for attention and semantic alignment as:
 
\begin{align}\label{update_z}
    \mathcal{L} = \mathcal{L}_{img} + \alpha_t\cdot \mathcal{L}_{att} , \quad
    z_t \leftarrow z_t- \nabla_{z_t} \mathcal{L} \odot mask .
\end{align}


To further incorporate semantic guidance, we adapt $c$ to the finer guidance $c'$ by computing their similarity and optimize the prompt embedding as:
\begin{align}
    \mathcal{L}_{prompt} = 1.0 - \operatorname{cosine}(\tau(c),  \tau(c')) ,
\end{align}

\vspace{-0.8cm}

\begin{align}\label{update_p}
    \tau(c) \leftarrow \tau(c)-\nabla_{\tau(c)} (\mathcal{L}_{prompt} + \mathcal{L}_{img}).
\end{align}

This process enriches the semantic information within the original prompt embedding $\tau(c)$ to guide the generation process effectively.
In our framework, we optimize this joint loss in the final $30$ of the total inference steps.
This integrated guidance not only strengthens semantic consistency but also mitigates unrealistic artifacts often encountered with standalone attention-based optimization, enabling more precise, contextually rich anomaly generation aligned with detailed prompt specifications.





With our proposed \method{} framework, we achieve promptable unseen anomaly generation. Notably, since \method{} is not trained on specific datasets, it is unrestricted by the distribution of available normal samples, making it applicable across a broad range of unseen object categories and anomaly types. 
Additionally, the attention maps define a probability distribution over each text token, allowing us to use the final smoothed attention map $\Bar{A}_0^j$ at time step $0$ as pixel-level annotation to localize the anomaly described by $c_j$.
\begin{figure}[t]
    \centering
    \includegraphics[width=0.9\linewidth]{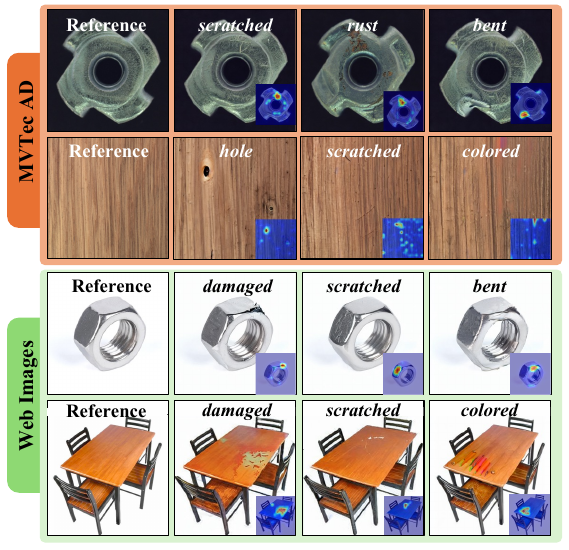}
    \vspace{-0.2cm}\caption{\textbf{Anomaly generation results for arbitrary objects and anomaly descriptions.} The bottom right presents the attention maps of the anomaly tokens.}
    \vspace{-0.4cm}
    \label{fig:ad_gen_diversity}
\end{figure}

\section{Experiment}





\subsection{Experimental Setup}\label{setup}

\textbf{Datasets.} We conduct extensive experiments on MVTec AD \cite{bergmann2019mvtec} and VisA \cite{zou2022spot}, two benchmarks for industrial anomaly detection. MVTec AD includes $5, 000$ images across $15$ object and texture categories with pixel-level annotations for defects, while VisA contains  $10, 821$ images from $12$ categories with annotations.

\noindent\textbf{Evaluation Metrics.} To evaluate the anomaly generation results quantitatively, we employ Inception Score (IS) to measure the generation quality and Intra-cluster pairwise LPIPS distance (IC-LPIPS)~\cite{ojha2021few} to evaluate the generation diversity.
We also validate the effectiveness of our generated image samples by training anomaly detection frameworks using our generated data. For the anomaly detection task, we employ five metrics to thoroughly evaluate the detection performance: image-level and pixel-level Area Under the Receiver Operating Characteristic (AUROC), denoted as I-AUC and P-AUC, respectively, the image-level and pixel-level max-F1 scores~\cite{WinClip}, denoted as I-F1 and P-F1, and Per-Region-Overlap~\cite{MVTec-AD} denoted as PRO.

\noindent\textbf{Implementation Details.}
For anomaly generation, we deploy the pretrained SD model
with $T=100$ and $\gamma=0.25$. Following ~\cite{zhang2024realnet}, we apply binary thresholding to extract a foreground mask for each normal guidance image, which acts as mask guidance for generation. More implementation details are provided in the supplementary material.

\begin{table}[t]
\centering
\fontsize{8pt}{8pt}\selectfont
\setlength{\tabcolsep}{2pt}
\caption{\textbf{Comparisons between anomaly generation quality and diversity with different anomaly generation methods.} 
AnoDiff: AnomalyDiffusion. }
\label{tab:is}
\renewcommand{\arraystretch}{1.2}
\resizebox{1.0\linewidth}{!}{%
\begin{tabular}{c|cc|cc|cc|cc}
\toprule[1.2pt]
 
\multirow{2}{*}{\textbf{Category}}  & \multicolumn{2}{c|}{\textbf{NSA~\cite{schluter2022natural}}} & \multicolumn{2}{c|}{\textbf{RealNet~\cite{zhang2024realnet}}} & \multicolumn{2}{c|}{\textbf{ AnoDiff~\cite{hu2024anomalydiffusion}}} & \multicolumn{2}{c}{\textbf{Ours}} \\
\cmidrule{2-9}
 & \textbf{IS $\uparrow$} & \textbf{IC-L $\uparrow$} & \textbf{IS $\uparrow$} & \textbf{IC-L $\uparrow$} & \textbf{IS $\uparrow$} & \textbf{IC-L $\uparrow$} & \textbf{IS $\uparrow$} & \textbf{IC-L $\uparrow$} \\
\midrule
                bottle & 1.35 & 0.11 & 1.58 & 0.16 & 1.58 & 0.19 & 1.73 & 0.17 \\ 
        cable & 1.55 & 0.38 & 1.65 & 0.41 & 2.13 & 0.41 & 2.06 & 0.41 \\ 
        capsule & 1.38 & 0.17 & 1.62 & 0.18 & 1.59 & 0.21 & 2.16 & 0.23 \\ 
        carpet & 1.03 & 0.22 & 1.03 & 0.25 & 1.16 & 0.24 & 1.10 & 0.34 \\ 
        grid & 2.26 & 0.36 & 2.24 & 0.37 & 2.04 & 0.44 & 2.31 & 0.38 \\ 
        hazelnut & 2.04 & 0.31 & 2.21 & 0.31 & 2.13 & 0.31 & 2.55 & 0.32 \\ 
        leather & 1.07 & 0.26 & 1.67 & 0.36 & 1.94 & 0.41 & 2.26 & 0.41 \\ 
        metal nut & 1.66 & 0.23 & 1.67 & 0.24 & 1.96 & 0.30 & 1.82 & 0.27 \\ 
        pill & 1.34 & 0.24 & 1.45 & 0.26 & 1.61 & 0.26 & 2.91 & 0.30 \\ 
        screw & 1.20 & 0.31 & 1.20 & 0.31 & 1.28 & 0.30 & 1.33 & 0.32 \\ 
        tile & 1.31 & 0.37 & 1.57 & 0.48 & 2.54 & 0.55 & 2.66 & 0.53 \\ 
        toothbrush & 1.18 & 0.17 & 1.19 & 0.18 & 1.68 & 0.21 & 1.64 & 0.22 \\ 
        transistor & 1.34 & 0.22 & 1.49 & 0.30 & 1.57 & 0.34 & 1.66 & 0.28 \\ 
        wood & 1.30 & 0.32 & 2.22 & 0.41 & 2.33 & 0.37 & 1.93 & 0.41 \\ 
        zipper & 1.52 & 0.22 & 1.88 & 0.24 & 1.39 & 0.25 & 2.14 & 0.33 \\ 
        \hline
        average & 1.44 & 0.26 & 1.64 & 0.30 & {\ul 1.80} & {\ul 0.32} & \textbf{2.02} & \textbf{0.33} \\ 
\bottomrule[1.2pt]
\end{tabular}
}
    \vspace{-0.2cm}
\end{table}

\begin{table*}[h]
\centering
\fontsize{8pt}{8pt}\selectfont
\setlength{\tabcolsep}{8pt}
\caption{\textbf{Comparison of 1-shot anomaly detection on MVTec AD and VisA.} Results are reported over 5 runs. The best results are in \textbf{bold}, and the second-best results are {\ul underlined}. }
\label{tab:few-shot}
\renewcommand{\arraystretch}{1.0}
\resizebox{\textwidth}{!}{%
\begin{tabular}{c|ccccc|ccccc}
\toprule[1.2pt]
  Metrics~$\rightarrow$ &
  \multicolumn{5}{c|}{MVTec AD} &
  \multicolumn{5}{c}{VisA} \\ \cmidrule(l){2-11} 
  Methods~$\downarrow$ &
  I-AUC &
  I-F1 &
  P-AUC &
  P-F1 &
  PRO &
  I-AUC &
  I-F1 &
  P-AUC &
  P-F1 &
  PRO \\ \midrule
  PaDiM~\cite{padim} &
  76.6\scriptsize{±3.1} &
  88.2\scriptsize{±1.1} &
  89.3\scriptsize{±0.9} &
  40.2\scriptsize{±2.1} &
  73.3\scriptsize{±2.0} &
  62.8\scriptsize{±5.4} &
  75.3\scriptsize{±1.2} &
  89.9\scriptsize{±0.8} &
  17.4\scriptsize{±1.7} &
  64.3\scriptsize{±2.4} \\
  PatchCore~\cite{Patchcore} &
  83.4\scriptsize{±3.0} &
  90.5\scriptsize{±1.5} &
  92.0\scriptsize{±1.0} &
  50.4\scriptsize{±2.1} &
  79.7\scriptsize{±2.0} &
  79.9\scriptsize{±2.9} &
  81.7\scriptsize{±1.6} &
  95.4\scriptsize{±0.6} &
  38.0\scriptsize{±1.9} &
  80.5\scriptsize{±2.5} \\
  WinCLIP+~\cite{WinClip} &
  93.1\scriptsize{±2.0} &
  {\ul 93.7\scriptsize{±1.1}} &
  95.2\scriptsize{±0.5} &
  {\ul 55.9\scriptsize{±2.7}} &
  {\ul 87.1\scriptsize{±1.2}} &
  83.8\scriptsize{±4.0} &
  {\ul 83.1\scriptsize{±1.7}} &
  96.4\scriptsize{±0.4} &
  {\ul 41.3\scriptsize{±2.3}} &
  {\ul 85.1\scriptsize{±2.1}} \\
  AnomalyGPT~\cite{anomalygpt} &
  94.1\scriptsize{±1.1} &
  - &
  95.3\scriptsize{±0.1} &
  - &
  - &
  {\ul 87.4\scriptsize{±0.8}} &
  - &
  96.2\scriptsize{±0.1} &
  - &
  - \\
  PromptAD~\cite{promptad} &
  {\ul 94.6\scriptsize{±1.7}} &
  - &
  \textbf{95.9\scriptsize{±0.5}} &
  - &
  - &
  86.9\scriptsize{±2.3} &
  - &
  {\ul 96.7\scriptsize{±0.4}} &
  - &
  - \\ \midrule
  \rowcolor{blue!8}\textbf{\method{} (Ours)} &
  \textbf{94.9\scriptsize{±0.4}} &
  \textbf{94.7\scriptsize{±0.4}} &
  {\ul 95.4\scriptsize{±0.2}} &
  \textbf{57.3\scriptsize{±0.0}} &
  \textbf{91.9\scriptsize{±0.0}} &
  \textbf{89.7\scriptsize{±0.8}} &
  \textbf{85.8\scriptsize{±0.5}} &
  \textbf{97.7\scriptsize{±0.4}} &
  \textbf{43.2\scriptsize{±0.4}} &
  \textbf{92.5\scriptsize{±0.1}} \\ 
  \bottomrule[1.2pt]
\end{tabular}
 
}
\end{table*}
\begin{table*}[h]
\centering
\fontsize{8pt}{8pt}\selectfont
\setlength{\tabcolsep}{8pt}
\caption{\textbf{Comparisons between 1-shot anomaly detection performance with different anomaly generation methods.} Since AnomalyDiffusion utilizes anomalous data for training and results in data leakage, we exclude it from ranking. Results are reported over 5 runs. The best results are in \textbf{bold}, and the second-best results are {\ul underlined}.  }
\label{tab:ad_generative}
\renewcommand{\arraystretch}{1.0}
\resizebox{\textwidth}{!}{%
\begin{tabular}{c|ccccc|ccccc}
\toprule[1.2pt]
 
   Metrics~$\rightarrow$ &
  \multicolumn{5}{c|}{MVTec AD} &
  \multicolumn{5}{c}{VisA} \\ \cmidrule(l){2-11} 
  Methods~$\downarrow$  &
  I-AUC &
  I-F1 &
  P-AUC &
  P-F1 &
  PRO &
  I-AUC &
  I-F1 &
  P-AUC &
  P-F1 &
  PRO \\ \midrule
\rowcolor[HTML]{F2F2F2} 
\small{AnomalyDiffusion~\cite{hu2024anomalydiffusion}} &
  94.4\scriptsize{±0.3} &
  94.4\scriptsize{±0.2} &
  95.3\scriptsize{±0.5} &
  57.3\scriptsize{±3.0} &
  92.2\scriptsize{±1.0} &
  - &
  - &
  - &
  - &
  - \\
DRAEM~\cite{zavrtanik2021draem} &
  93.6\scriptsize{±0.3} &
  {\ul 94.2\scriptsize{±0.4}} &
  {\ul95.1\scriptsize{±0.1}} &
  56.0\scriptsize{±0.9} &
  {\ul91.8\scriptsize{±0.1}} &
  86.0\scriptsize{±0.7} &
  83.0\scriptsize{±0.9} &
  {\ul 97.5\scriptsize{±0.1}} &
  {\ul 42.6\scriptsize{±0.7}} &
  {\ul 92.6\scriptsize{±0.6}} \\
NSA~\cite{schluter2022natural} &
  {\ul 94.0\scriptsize{±0.5}} &
  {\ul 94.2\scriptsize{±0.3}} &
  {\ul 95.1\scriptsize{±0.1}} &
  56.1\scriptsize{±0.5} &
  {\ul 91.8\scriptsize{±0.2}} &
  {\ul86.2\scriptsize{±2.0}} &
  {\ul83.1\scriptsize{±1.2}} &
  97.4\scriptsize{±0.1} &
  40.8\scriptsize{±0.5} &
  92.3\scriptsize{±0.3} \\
RealNet~\cite{zhang2024realnet} &
  92.7\scriptsize{±0.7} &
  93.6\scriptsize{±0.3} &
  {\ul95.1\scriptsize{±0.1}} &
  {\ul56.3\scriptsize{±1.3}} &
  91.7\scriptsize{±0.1} &
  86.0\scriptsize{±1.4} &
  82.9\scriptsize{±1.1} &
  {\ul 97.5\scriptsize{±0.2}} &
  41.9\scriptsize{±1.8} &
  \textbf{92.8\scriptsize{±0.3}} \\ \midrule
\rowcolor{blue!8}\textbf{\method{} (Ours)} & \textbf{94.9\scriptsize{±0.4}} & \textbf{94.7\scriptsize{±0.4}} & \textbf{95.4\scriptsize{±0.2}} & 
\textbf{57.3\scriptsize{±0.0}} & 
\textbf{91.9\scriptsize{±0.0}} & \textbf{89.7\scriptsize{±0.8}} & \textbf{85.8\scriptsize{±0.5}} & \textbf{97.7\scriptsize{±0.4}} & \textbf{43.2\scriptsize{±0.4}} & 92.5\scriptsize{±0.1} \\ \bottomrule[1.2pt]
\end{tabular}
}
\end{table*}

\subsection{Anomaly Generation Results}
\label{agresults}

\begin{figure}[t]
\centering
  \includegraphics[width=1.0\linewidth]{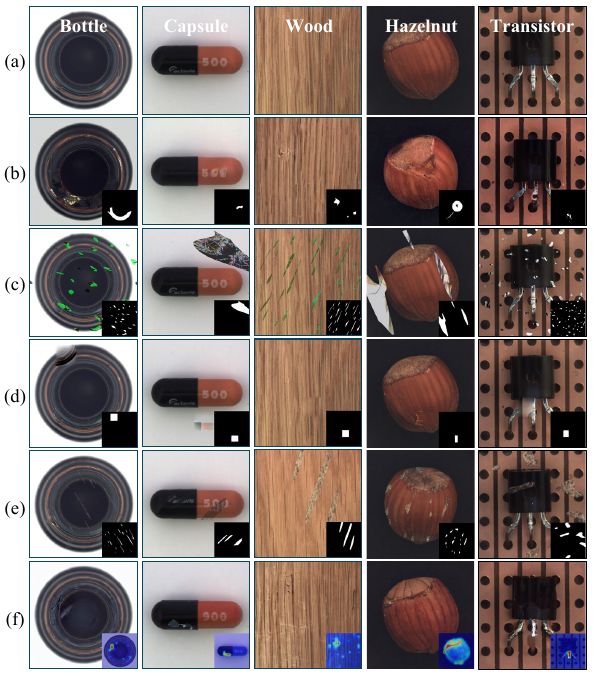}
    \vspace{-0.6cm}
    \caption{\textbf{Qualitative comparisons between existing anomaly generation methods.} 
    (a) Reference, (b) AnomalyDiffusion, (c) DREAM, (d) NSA, (e) RealNet, (f) Our \method{}. 
    }
    \vspace{-0.4cm}
    \label{fig:generation_comparison}
\end{figure}

\textbf{Unseen Anomaly Generation.}
In \cref{fig:ad_gen_diversity}, we present the generation results of \method{} on MVTec AD and web images. Compared to existing data generation methods like \cite{zhang2021defect, hu2024anomalydiffusion, hu2024anomalyxfusion}, our \method{} requires no training on normal or anomalous samples. Thus, \method{} is not constrained by the limited available training data distribution, presenting universal anomaly generation capabilities on unseen data and arbitrary anomaly types. 
This shows our method's great potential in personalized anomaly generation with arbitrary normal patterns and anomalous descriptions, and may greatly contribute to the enhancement of anomaly detection models for a specific category even without real collected anomalous samples.
Additional generation results are provided in the supplementary material.


\noindent\textbf{Comparison with other anomaly generation methods.} 
We compare the generation quality of \method{} on MVTec AD with several existing anomaly generation methods, including DRAEM~\cite{zavrtanik2021draem}, NSA \cite{schluter2022natural}, RealNet \cite{zhang2024realnet}, and AnomalyDiffusion \cite{hu2024anomalydiffusion}. Among these methods, AnomalyDiffusion trains on $1/3$ anomalous data from the test set. 
\cref{fig:generation_comparison} presents the generated anomalies of these methods. We include more examples in the supplementary material. 
The examples clearly show that DRAEM, NSA, and RealNet introduce inauthentic patterns, while
AnomalyDiffusion and \method{} can generate more realistic anomalous samples. Notably, AnomalyDiffusion and RealNet rely on available anomalous training data, while \method{} generates without training and generalizes to unseen object types and anomalies during inference.

We present quantitative evaluations of generation quality in \cref{tab:is}. For NSA, RealNet, and \method{}, we condition on MVTec AD training images to generate anomaly samples.
As DRAEM crops random textures from external datasets, we do not include it for comparison~\cite{hu2024anomalydiffusion}. 
The result demonstrates that our model generates anomaly data with both the highest quality and diversity.
Additionally, a user study on anomaly generation quality is included in the supplementary material, providing further evidence of the superior generation quality achieved by \method{}.



\subsection{Anomaly Detection Results}~\label{sec:anomaly_detection}
The key advantage of our method lies in its capability to generate unseen anomalies without needing any normal or abnormal samples for training, making it highly effective under scenarios with data scarcity. To further demonstrate the effectiveness of our method, we evaluate it under the challenging 1-shot anomaly detection scenario, where only one single normal sample is accessible and no abnormal samples are available.
Following the method proposed in ~\cite{anomalygpt}, we condition on a single normal image to generate $100$ random anomalous samples. We use the generated samples with the attention map of the anomaly token $c_j$ as the corresponding anomaly mask to facilitate anomaly detection training.
Further details on the generation process and anomaly detection training are provided in the supplementary material.
\cref{tab:few-shot} presents the comparison results between our \method{} and existing few-shot AD methods, including two full-shot methods ~\cite{padim, Patchcore} in the few-shot settings, and three CLIP-based few-shot methods ~\cite{WinClip, anomalygpt, promptad}.
As can be seen, our generated anomalous samples consistently outperform other alternatives in the majority of the metrics.

To further compare with other anomaly generation methods under the data scarcity scenarios, we follow the same 1-shot setup and substitute the synthetic training data with anomalous samples generated by alternative methods. For DRAEM, NSA, and RealNet, we condition on the same single normal sample as for our method and generate $100$ random anomalous samples for training. We report the results in \cref{tab:ad_generative}.
The results indicate that anomalous samples generated by \method{} yield the best detection results. Additionally, we compare with AnomalyDiffusion, which has seen all the training samples and some of the test anomalous samples and thus introduced data leakage. Despite not being trained with test set anomalous samples, our method produces comparable anomaly detection performance.

\subsection{Ablations}

\begin{figure}
    \centering
    \includegraphics[width=1.0\linewidth]{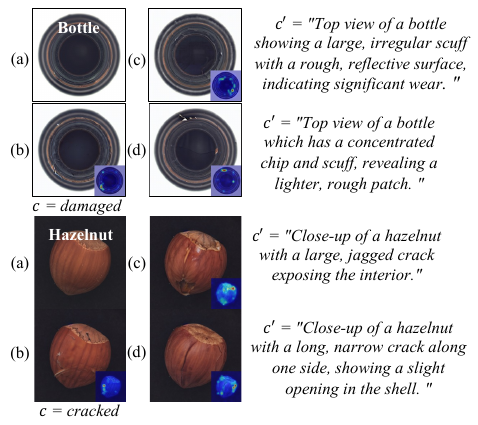}
    \vspace{-0.6cm}
    \caption{\textbf{Ablations on different types of prompt guidance.} (a) Normal image, (b) Generation results given anomaly type prompt $c$ and (c)(d) Generation results given $c$ and detailed prompt $c'$.}
    \vspace{-0.4cm}
    \label{fig:enter-label}
\end{figure}

\noindent\textbf{Ablation on different types of prompt guidance.}
We compare the generation results under different types of guidance in our \method{}.
The results in \cref{fig:enter-label} demonstrate that our method reflects the semantics of the anomaly prompts $c$ generated by GPT-4. By providing the detailed prompt description $c'$, we are able to get more fine-grained and realistic generation results with more diversity.



\begin{figure}
    \centering
    \includegraphics[width=1.0\linewidth]{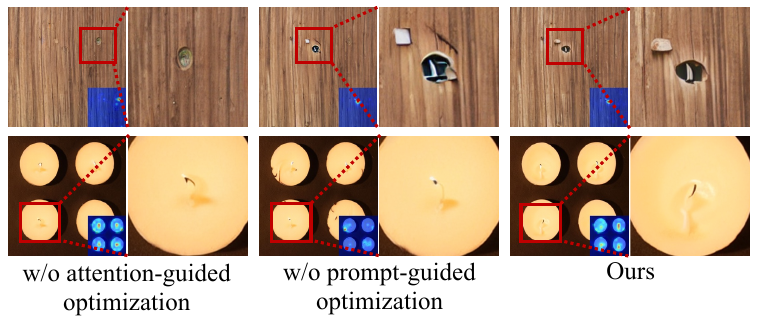}
    \vspace{-0.6cm}
    \caption{\textbf{Ablation on optimization strategies.} Anomaly types are \textit{"hole"} and \textit{"melted"} respectively.}
    \label{fig:module}
\end{figure}

\begin{figure}
    \centering
    \includegraphics[width=1.0\linewidth]{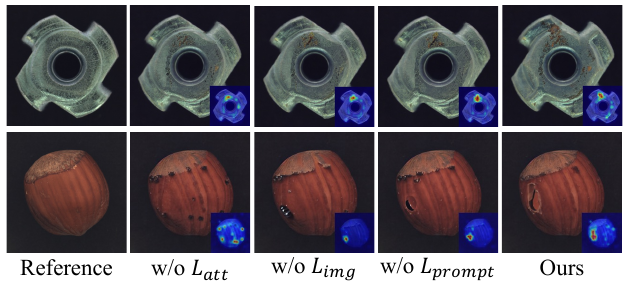}
    \vspace{-0.6cm}
    \caption{\textbf{Ablation on prompt-guided optimization objectives.} Anomaly types are \textit{"rust"} and \textit{"hole"} respectively.}
    \label{fig:loss}
\end{figure}

\noindent\textbf{Ablation on attention-guided \& prompt-guided optimization.}
We evaluate the effectiveness of attention-guided optimization and prompt-guided optimization separately, as illustrated in \cref{fig:attention}
(d-f), with additional results shown in~\cref{fig:module}. Our results show that attention guidance successfully enforces the generation of the specified anomaly description, ensuring it is not overlooked by SD. Furthermore, detailed prompt guidance enhances the realism of the generated samples.
We also analyze the impact of different loss terms in \cref{update_z} and \cref{update_p} for optimizing $z$ and $\tau(c)$ respectively, as shown in~\cref{fig:loss}. The attention loss term improves focus on the anomaly region, while prompt-based terms help enhance generation quality. Notably, when the anomaly type is unambiguous (e.g., \textit{"rust"}), prompt guidance has a reduced impact on the generation quality. 
Additional visual results from the ablation study are provided in the supplementary material.

\section{Conclusion}
In this paper, we propose \method{}, a novel framework that utilizes SD for unseen anomaly generation, allowing users to generate realistic anomalous samples from arbitrary normal images of objects and anomaly text descriptions. Our framework incorporates attention-guided anomaly optimization to direct SD's attention toward anomaly concepts and prompt-guided anomaly refinement to enhance the authenticity of generated samples.
Compared to existing methods, \method{} achieves authentic unseen anomaly generation without additional training. 
Extensive experimental results demonstrate the effectiveness of \method{} in generating high-quality anomalies that improve the performance of downstream anomaly detection tasks.

Currently, \method{} relies on normal image guidance and textual descriptions for anomaly generation. While text prompts offer flexibility, they can sometimes lack sufficient guidance due to  CLIP’s limited understanding of lengthy and specialized descriptions as well as the precision required for complex anomalies. Future work could address this by incorporating additional inputs, such as one-shot anomalous image. This would increase  \method{}'s versatility, enabling more precise anomaly generation across various applications.
With its demonstrated generation capabilities and ability to generalize to unseen anomalies, \method{} has significant potential to advance foundation models for anomaly detection.

\clearpage

{
    \small
    \bibliographystyle{ieeenat_fullname}
    \bibliography{main}
}

\clearpage
\appendix
\setcounter{page}{1}
\maketitlesupplementary

\section*{Overview}
The supplementary material presents the following sections to strengthen the main manuscript:

\begin{itemize}[label=---, left=1.5em]
    \item \textbf{Sec.~\ref{implementation}} shows more implementation details.
    \item \textbf{Sec.~\ref{appen:survey}} presents a user study on anomaly generation quality.
    \item \textbf{Sec.~\ref{append:more_generation_results}} presents more anomaly generation results.
    \item \textbf{Sec.~\ref{mask}} presents results on mask controllability.
    \item \textbf{Sec.~\ref{hype}} shows ablations on hyperparameters.
    \item \textbf{Sec.~\ref{append:more_ablations}} shows more ablations on attention-guided and prompt-guided optimization.
    \item \textbf{Sec.~\ref{append:more_detection}} shows more anomaly detection results.
\end{itemize}



\section{More Implementation Details}\label{implementation}
\textbf{Stable Diffusion} For our proposed \method, we set the inference steps to 100 and $\gamma=0.25$ with stable-diffusion-v1-5. For optimization, we set $\lambda=10$ and $\Delta t$ to $1.0/T$. 
As implemented in~\cite{chefer2023attend}, the maximum value threshold to stop the iterative optimization at one time step $t$ is set to ${ 0.05, 0.5, 0.8}$, increasing with the denoising diffusion process.
All experiments are run on a single NVIDIA A100-SXM4-80GB GPU.

\begin{figure*}[t]
    \centering
    \includegraphics[width=0.8\textwidth]{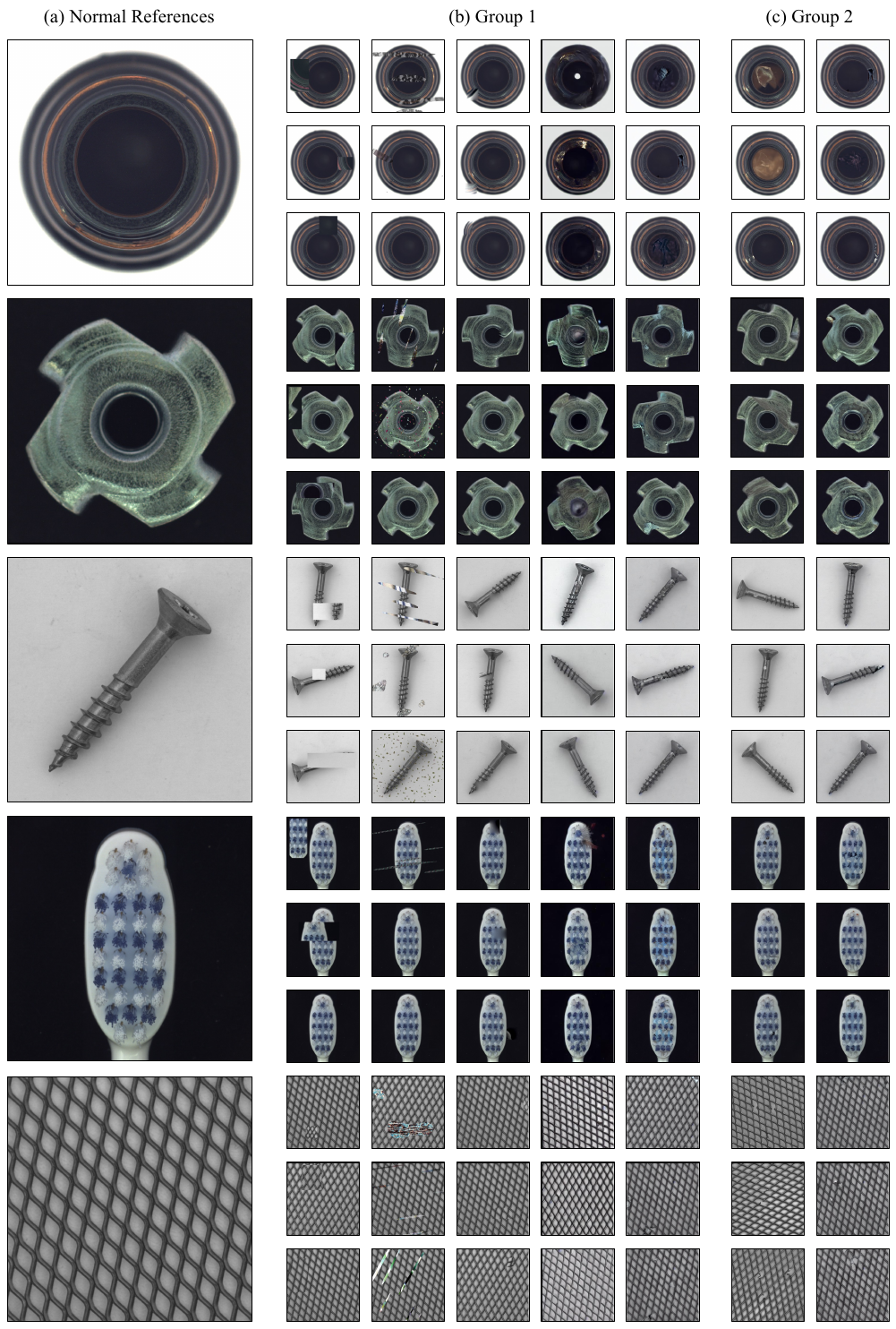}
    \caption{\textbf{Samples for user study.} (a) Normal references. (b) Group1: Samples of each column from left to right are generated by: Cut\&Paste; DRAEM; NSA; AnomalyDiffusion; the proposed \method; (c) Group2: Samples of each column from left to right are real anomalous data samples from the dataset and anomalous images generated by our method \method.}
    \label{fig:survey}
\end{figure*}

\noindent\textbf{Anomaly detection framework}\label{append:vlad}
We adapt the anomaly detection framework proposed in AnomalyGPT~\cite{anomalygpt}, which deploys CLIP~\cite{clip} to compute the vision-language and vision-vision similarities and aggregate these similarities for anomaly detection. 
The similarity between visual tokens and text embeddings for normal/anomalous states can indicate the abnormal level of visual tokens~\cite{WinClip}. Specifically, for a given image, we first extract its patch tokens $\mathbf{F}^i_{patch} \in \mathbb{R}^{H_i \times W_i \times C_i}$ and image token $\mathbf{F}_{image} \in \mathbb{R}^{1 \times C}$ using the CLIP visual encoder, where $i$ indicates the tokens that are extracted from the $i$-th stage of the image encoder. Then, text embeddings $\mathbf{F}_{text} \in \mathbb{R}^{2 \times C}$ representing normal/abnormal states are extracted via the CLIP text encoder. Since the extracted patch tokens have not undergone the final image-text alignment and cannot be directly compared with text features, we use a lightweight feature adapter comprising only a linear layer to project patch tokens for both fine-tuning and dimension alignment between visual and text embeddings, producing $\hat{\mathbf{F}}^i_{patch} \in \mathbb{R}^{H_i \times W_i \times C}$. The detection and localization results based on vision-language similarity can then be obtained as follows:
\begin{align}
   S_{VL}&=\text{softmax}(\mathbf{F}_{image} \cdot \mathbf{F}_{text}^{T}),\\
M_{VL}&=\text{Upsample}\left(\sum_{i\in{\mathcal{H}}}{\text{softmax}(\hat{\mathbf{F}}^i_{patch} \cdot \mathbf{F}_{text}^{T})}\right) ,
\end{align}
\noindent where $\mathcal{H}$ is the list of selected stages, and $S_{VL}$ and $M_{VL}$ denote the image- and pixel-level anomaly scores, respectively.
When some normal samples are available, we utilize the same visual encoder and feature adapter to extract multi-hierarchy normal patch tokens and store them in memory banks $\mathbf{B}^i \in \mathbb{R}^{N_i \times C}$. Then, for the testing patch tokens, we compute the distance between each token and its most similar counterpart in the memory bank, and the localization result $M_{VV}$ based on vision-vision similarities is yielded as follows:
\begin{equation}
M_{VV}=\text{Upsample}\left(\sum_{i\in{\mathcal{H}}}{(1-\max(\hat{\mathbf{F}}^i_{patch} \cdot (\mathbf{B}^i)^{T}))}\right)
\end{equation}
\noindent The maximum value of $M_{VV}$ is taken as the image-level anomaly scores $S_{VV}$. The predictions from vision-language and vision-vision similarities are summed up as final predictions.
We train the anomaly detection model on the available normal sample and the synthetic samples for 200 epochs with batch size 16. We use Adam optimizer with a learning rate of 1e4 and the CosineAnnealingLR scheduler. 

\section{User Study on Anomaly Generation Quality}\label{appen:survey}
To better assess the quality of our generated anomalous samples, we conducted a user study with 20 participants. The participants were shown exemplar normal samples of the five tested categories and asked to choose the most realistic anomalous images.
We provided the participants with two groups of samples, as shown in Figure~\ref{fig:survey}. In group 1, we randomly sampled three images each from 100 anomalous samples generated by Cut\&Paste, DRAEM, NSA, AnomalyDiffusion, and our proposed \method. Participants were asked to choose the three most realistic images for each category. For this group, we get a total vote of 300:  $20(participants)\cdot 5(categories) \cdot 3 (selected \, samples \, per \, category)$. In group 2, we randomly sampled two images each from real anomalies in the test set and from the 100 images generated by our method. Participants were then asked to choose the two most realistic images for each category. For this group, we get a total vote of 200: $20(participants)\cdot 5(categories) \cdot 2 (selected \, samples \, per \, category)$.
The survey results are reported in Table~\ref{tab:methods_comparison} where we show the total votes from the 20 participants for each group.  It is evident that our method surpasses other anomaly generation methods in terms of authenticity, even compared to AnomalyDiffusion, which leverages real test samples for training. Additionally, when mixed with real normal samples, our generated anomalous images are realistic enough to be misclassified.

\begin{table}[h]
\centering
\caption{Results of \textbf{user study} for anomaly generation quality assessment across two groups.}
\label{tab:methods_comparison}
\resizebox{\linewidth}{!}{%
\begin{tabular}{ccccccc}
\toprule[1.5pt]
\multicolumn{5}{c}{Group 1}                                 & \multicolumn{2}{c}{Group 2} \\ \cmidrule(lr){1-5} \cmidrule(lr){6-7}
Cut\&Paste & DRAEM & NSA & AnomalyDiffusion & \cellcolor{blue!10}{\textbf{Ours}} & Real    & \cellcolor{blue!10}{\textbf{Ours}}    \\ \cmidrule(lr){1-5} \cmidrule(lr){6-7}
6          & 33    & 61  & 64               & \cellcolor{blue!10}{136}           & 99      & \cellcolor{blue!10}{101}              \\ \bottomrule[1.5pt]
\end{tabular}
}
\end{table}

\begin{table*}[t]
\centering
\caption{\textbf{Corresponding descriptions for anomaly generation} in Figure~\ref{fig:universal_generation1}}
\label{tab:descriptions}
\resizebox{\linewidth}{!}{%
\begin{tabular}{lll}
\toprule[1.2pt]
$\text{[CLS]}$                    & $\text{[Anomaly State]}$   & GPT-Generated Detailed Descriptions                                                      \\ \midrule
\multirow{5}{*}{Screw}   & Damaged           & The head of the screw is damaged or worn-out.                                            \\
                         & Scratched         & A screw with several deep, thin scratches.                                               \\
                         & Broken on the top & The screw is partially broken on the tip.                                                \\
                         & Bent              & The screw is bent along its shaft, making it difficult to use.                           \\
                         & Rust              & The screw is covered in rust, weakening its structure and making it   difficult to use.  \\ \midrule
\multirow{5}{*}{Leather} & Damaged           & Leather that has dried out or aged can develop damaged cracks.                           \\
                         & Scratched         & Leather that has been scratched with deep, gouging lines.                                \\
                         & Cut               & The leather has a visible cut, creating a split in its surface.                          \\
                         & Stained           & Leather that is discolorations caused by spills (oil, ink, or dyes).                     \\
                              & Wet       & Leather that appears darker, unevenly discolored, and may develop a tacky   texture or water spots as it dries. \\ \midrule
\multirow{5}{*}{Bowl}    & Broken            & A broken bowl that has visible cracks with jagged or uneven edges.                       \\
                         & Dirty             & A dirty bowl that has visible stains on its surface.                                     \\
                         & Colored           & The bowl has uneven patches of color, giving it a stained or discolored   appearance.    \\
                              & Cracked   & The bowl has a visible crack running through its surface, compromising   its strength and usability.            \\
                              & Stained   & The bowl has visible stains, with patches of discoloration or residue   marring its surface.                    \\ \midrule
\multirow{5}{*}{Vase}    & Broken            & The vase has fragments missing, making it unusable.                                      \\
                              & Cracked   & The vase has a visible crack running along its surface, threatening its   structural integrity.                 \\
                         & Colored           & The vase has patches of uneven hues that alter its original appearance.                  \\
                         & Hole              & The vase has a hole piercing its surface, preventing it from holding   liquids properly. \\
                         & Deformed          & The vase is misshapen, with uneven curves.                                               \\ \midrule
\multirow{5}{*}{Phone Screen} & Scratched & The phone screen has visible scratches, with fine lines marring its   smooth surface.                           \\
                         & Broken            & The phone screen is shattered, with pieces of glass splintered or   missing.             \\
                         & Cracked           & The phone screen has a crack running across it                                           \\
                         & Colored           & The phone screen displays abnormal patches of color, such as rainbow   streaks.          \\
                         & Fingerprint       & The phone screen has a smudged fingerprint, leaving an oily mark on its   surface.       \\ \bottomrule[1.2pt]
\end{tabular}
}
\end{table*}

\section{More Anomaly Generation Results}~\label{append:more_generation_results}

\begin{figure}[h]
    \centering
    \includegraphics[width=1.0\linewidth]{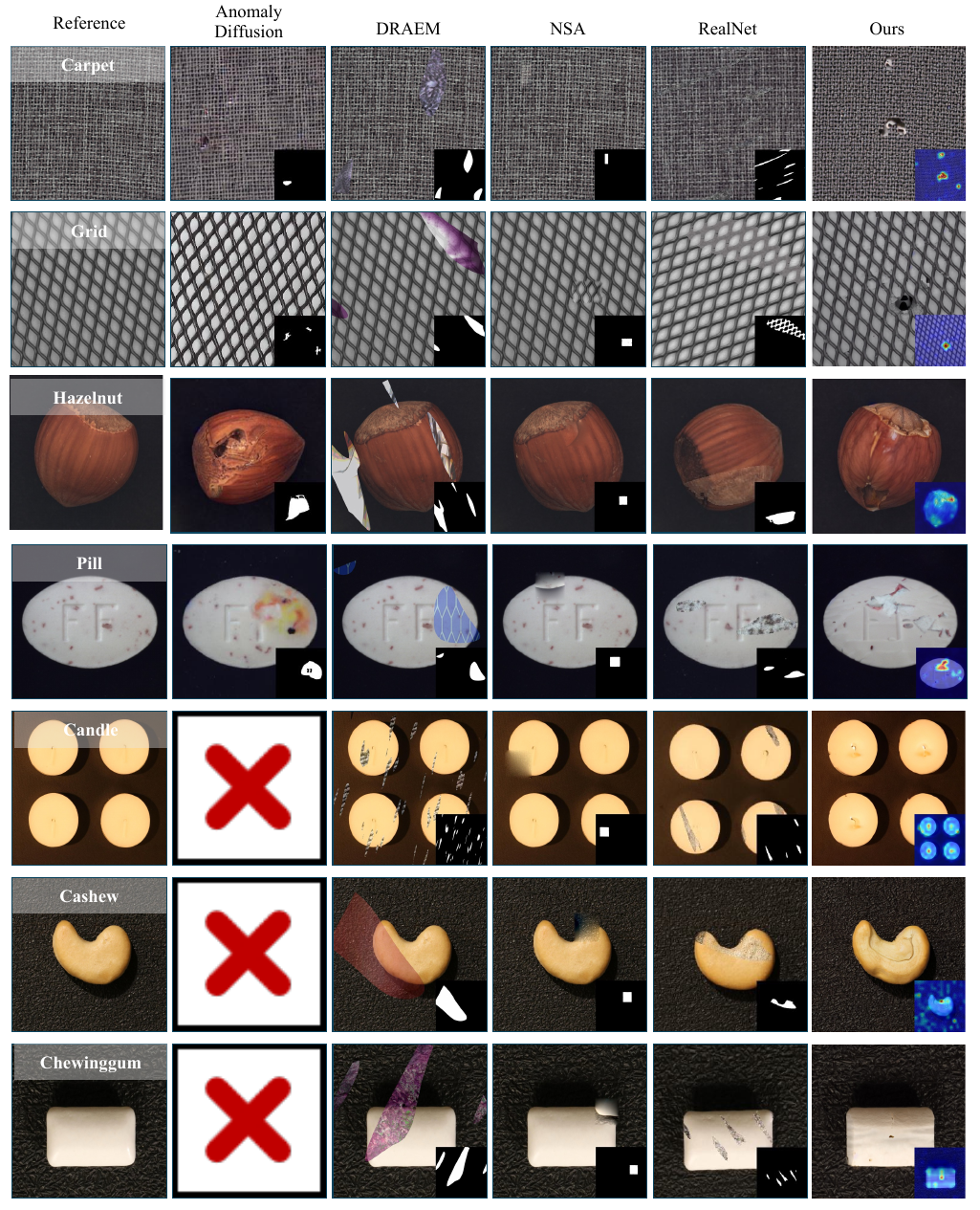}
    \caption{\textbf{Qualitative comparisons between existing anomaly generation methods.} Since AnomalyDiffusion does not provide results on VisA, its corresponding generation results are replaced by a blank.}
    \label{fig:generation_comparison_appendix}
\end{figure}

\begin{figure}[h]
    \centering
    \includegraphics[width=\linewidth]{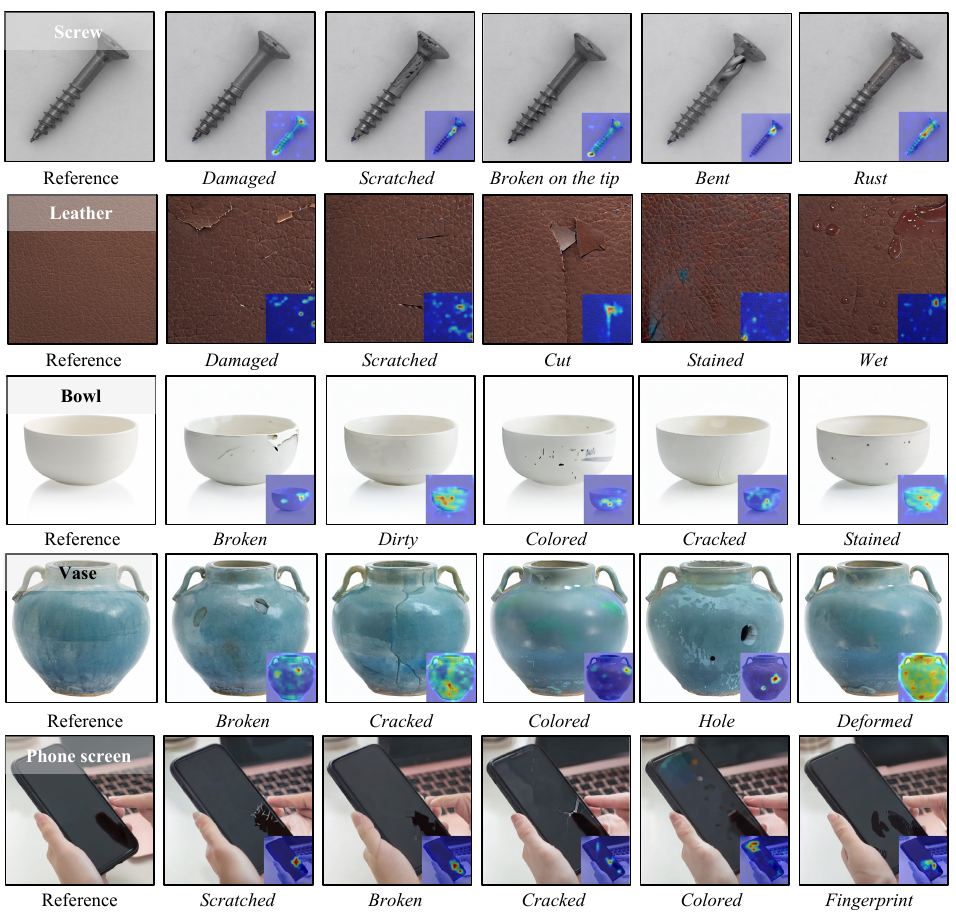}
    \caption{\textbf{Anomaly generation results} for arbitrary objects and anomaly descriptions.}
    \label{fig:universal_generation1}
\end{figure}

More comparisons of anomaly generation results between other anomaly generation methods and \method{} are provided in Figure \ref{fig:generation_comparison_appendix}. Additionally, to show the generalization ability and controllability of \method, Figure \ref{fig:universal_generation1} show more generation results with different object types and anomaly types with descriptions provided in Table~\ref{tab:descriptions}.

\section{Ablations on Mask Control}\label{mask}
\begin{figure}[h]
    \centering
    \includegraphics[width=0.4\textwidth]{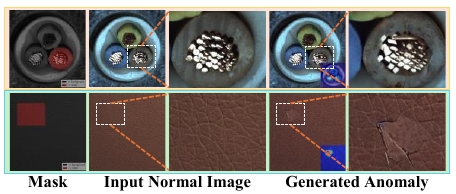}
    \caption{\textbf{Ablation on mask control.}}
    \label{fig:mask}
\end{figure}

In Figure~\ref{fig:mask}, we show results using manually labeled regional masks for more precise anomaly locations, demonstrating our finer controllability over anomaly regions when needed.

\section{Ablations on Hyperparameters}\label{hype}
\begin{figure}[h]
    \centering
    \includegraphics[width=0.4\textwidth]{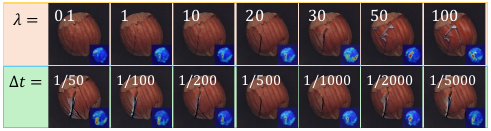}
    \caption{\textbf{Ablation on hyperparameters.}}
    \label{fig:hyp}
\end{figure}

In Figure~\ref{fig:hyp}, we present the visual results ablations on the hyperparameters $\lambda$ and $\Delta t$ in Equation (9).
It shows that a large $\lambda$ value causes artifacts due to overly fast updates, while a small $\lambda$ value results in insufficient optimization. A large $\Delta t$ introduces artifacts from fast updates early in the process, while a small $\Delta t$ leads to artifacts from inadequate detail refinement in the final steps. 

\section{More Ablations on attention-guided \& prompt-guided optimization}~\label{append:more_ablations}
\begin{figure*}[h]
    \centering
    \includegraphics[width=1.0\textwidth]{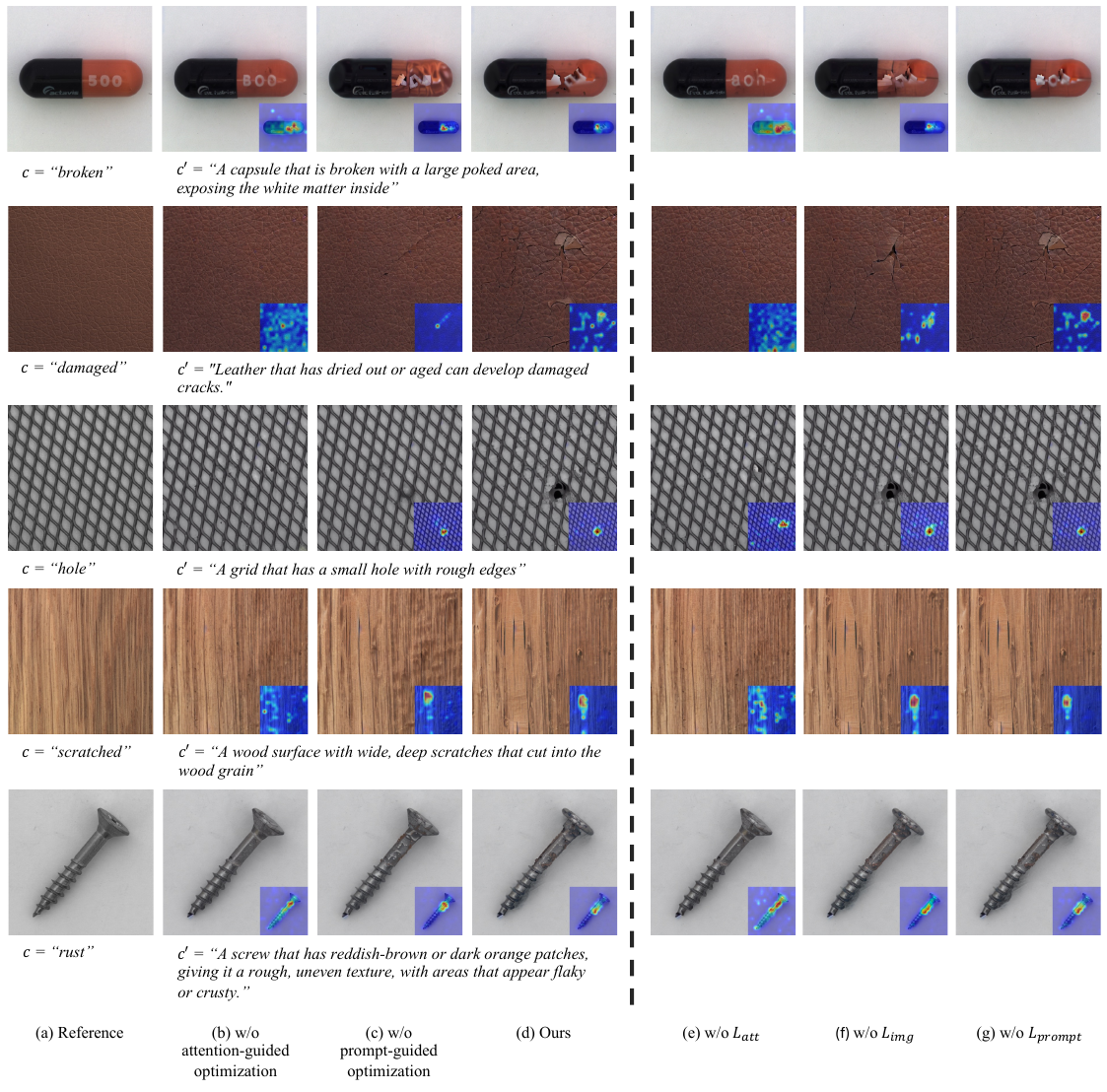}
    \caption{\textbf{Ablation on optimization strategies and objectives.}}
    \label{fig:module_appendix}
\end{figure*}

In Figure~\ref{fig:module_appendix}, we provide more results from ablations on different optimization strategies, as shown in (b)(c)(d), and on prompt-guided optimization objectives, as shown in (d)(e)(f)(g). The results from ablations on different optimization strategies further validate that our attention guidance module effectively enforces the generation of specified anomalies, while the detailed prompt guidance module enhances semantic richness and improves generation quality.

For the prompt-guided optimization objectives, the best overall results are achieved with our proposed method, which incorporates all optimization objectives. 
We observe that in some cases (as shown in the first two rows for capsule and leather examples), the image and prompt optimization objectives significantly improve generation results for challenging concepts. In other cases, the difference is less pronounced. 
Across all scenarios, $L_{img}$ consistently helps generate more salient anomaly patterns in the image with more concentrated anomaly attention maps. Meanwhile, in cases where the anomaly description is less ambiguous (e.g., "rust" or "hole"), the impact of $L_{prompt}$ on optimization is relatively minor.

\section{More Anomaly Detection Results}~\label{append:more_detection}
\begin{figure*}[h]
    \centering
    \includegraphics[width=0.9\textwidth]{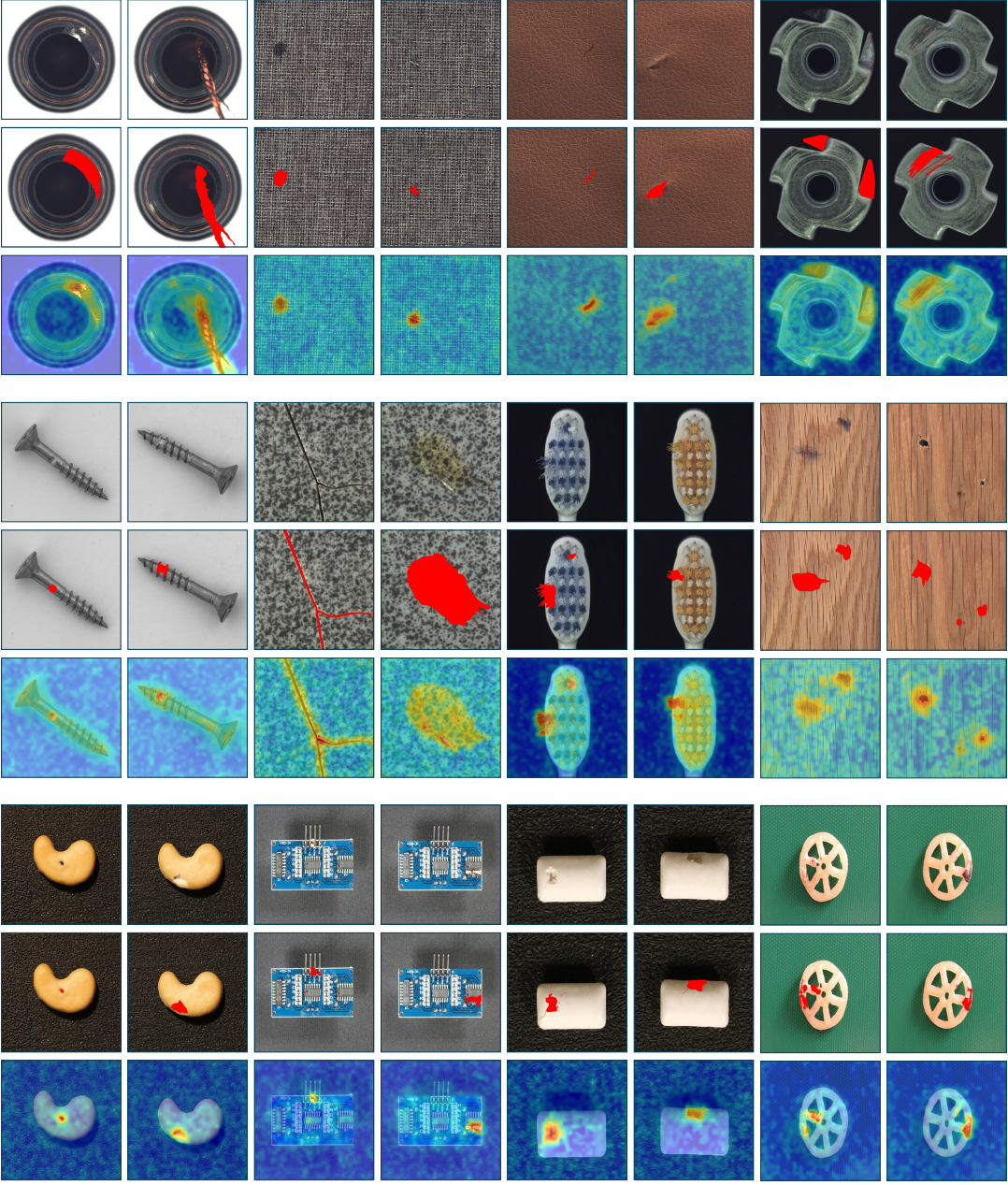}
    \caption{\textbf{Anomaly detection results in the 4-shot setup.} For each pair, the original image, ground truth, and detection results are listed from the top to the bottom.}
    \label{fig:anomaly_visualization_appendix}
\end{figure*}

In this section, we present additional results in Table~\ref{tab:few-shot-supp} for 2-shot and 4-shot anomaly detection. Specifically, we condition on 2 and 4 normal images, generating 200 and 400 synthetic anomalous images respectively for training. Additionally,  we provide per-category few-shot anomaly detection results on MVTec and VisA in Table~\ref{tab:mvtec_1shot} to Table~\ref{tab:visa_4shot}.
Visual results of few-shot anomaly detection are provided in Figure \ref{fig:anomaly_visualization_appendix}. Full-shot detection results are listed in Table~\ref{tab:full_shot} with per-category results in Table~\ref{tab:full_shot_mvtec_per_category} and Table~\ref{tab:full_shot_visa_per_category}. Specifically, we condition on all normal images, generating 3-5 anomalous images with each normal image for training.
We provide t-test results of few-shot anomaly detection in Table~\ref{tab:ttest} for statistical completeness.

\begin{table*}[t]
\centering
\caption{\textbf{Comparison of few-shot anomaly detection on MVTec AD and VisA.} Results are reported over 5 runs. The best results are in \textbf{bold}, and the second-best results are {\ul underlined}. }
\label{tab:few-shot-supp}
\renewcommand{\arraystretch}{1.0}
\resizebox{1.0\textwidth}{!}{%
\begin{tabular}{@{}c|c|ccccc|ccccc@{}}
\toprule[1.5pt]
\multirow{2}{*}{Setup} &
  \multirow{2}{*}{Methods} &
  \multicolumn{5}{c|}{MVTec AD} &
  \multicolumn{5}{c}{VisA} \\ \cmidrule(l){3-12} 
 &
   &
  I-AUC &
  I-F1 &
  P-AUC &
  P-F1 &
  PRO &
  I-AUC &
  I-F1 &
  P-AUC &
  P-F1 &
  PRO \\ \midrule
\multirow{6}{*}{1-shot} &
  PaDiM &
  76.6\scriptsize{±3.1} &
  88.2\scriptsize{±1.1} &
  89.3\scriptsize{±0.9} &
  40.2\scriptsize{±2.1} &
  73.3\scriptsize{±2.0} &
  62.8\scriptsize{±5.4} &
  75.3\scriptsize{±1.2} &
  89.9\scriptsize{±0.8} &
  17.4\scriptsize{±1.7} &
  64.3\scriptsize{±2.4} \\
 &
  PatchCore &
  83.4\scriptsize{±3.0} &
  90.5\scriptsize{±1.5} &
  92.0\scriptsize{±1.0} &
  50.4\scriptsize{±2.1} &
  79.7\scriptsize{±2.0} &
  79.9\scriptsize{±2.9} &
  81.7\scriptsize{±1.6} &
  95.4\scriptsize{±0.6} &
  38.0\scriptsize{±1.9} &
  80.5\scriptsize{±2.5} \\
 &
  WinCLIP+ &
  93.1\scriptsize{±2.0} &
  {\ul 93.7\scriptsize{±1.1}} &
  95.2\scriptsize{±0.5} &
  {\ul 55.9\scriptsize{±2.7}} &
  {\ul 87.1\scriptsize{±1.2}} &
  83.8\scriptsize{±4.0} &
  {\ul 83.1\scriptsize{±1.7}} &
  96.4\scriptsize{±0.4} &
  {\ul 41.3\scriptsize{±2.3}} &
  {\ul 85.1\scriptsize{±2.1}} \\
 &
  AnomalyGPT &
  94.1\scriptsize{±1.1} &
  - &
  95.3\scriptsize{±0.1} &
  - &
  - &
  {\ul 87.4\scriptsize{±0.8}} &
  - &
  96.2\scriptsize{±0.1} &
  - &
  - \\
 &
  PromptAD &
  {\ul 94.6\scriptsize{±1.7}} &
  - &
  \textbf{95.9\scriptsize{±0.5}} &
  - &
  - &
  86.9\scriptsize{±2.3} &
  - &
  {\ul 96.7\scriptsize{±0.4}} &
  - &
  - \\
 &
  \cellcolor{blue!10}\textbf{Ours} &
  \cellcolor{blue!10}\textbf{94.9\scriptsize{±0.4}} &
  \cellcolor{blue!10}\textbf{94.7\scriptsize{±0.4}} &
  \cellcolor{blue!10}{\ul 95.4\scriptsize{±0.2}} &
  \cellcolor{blue!10}\textbf{57.3\scriptsize{±0.0}} &
  \cellcolor{blue!10}\textbf{91.9\scriptsize{±0.0}} &
  \cellcolor{blue!10}\textbf{89.7\scriptsize{±0.8}} &
  \cellcolor{blue!10}\textbf{85.8\scriptsize{±0.5}} &
  \cellcolor{blue!10}\textbf{97.7\scriptsize{±0.4}} &
  \cellcolor{blue!10}\textbf{43.2\scriptsize{±0.4}} &
  \cellcolor{blue!10}\textbf{92.5\scriptsize{±0.1}} \\ \midrule
\multirow{6}{*}{2-shot} &
  PaDiM &
  78.9\scriptsize{±3.1} &
  89.2\scriptsize{±1.1} &
  91.3\scriptsize{±0.7} &
  43.7\scriptsize{±1.5} &
  78.2\scriptsize{±1.8} &
  67.4\scriptsize{±5.1} &
  75.7\scriptsize{±1.8} &
  92.0\scriptsize{±0.7} &
  21.1\scriptsize{±2.4} &
  70.1\scriptsize{±2.6} \\
 &
  PatchCore &
  86.3\scriptsize{±3.3} &
  92.0\scriptsize{±1.5} &
  93.3\scriptsize{±0.6} &
  53.0\scriptsize{±1.7} &
  82.3\scriptsize{±1.3} &
  81.6\scriptsize{±4.0} &
  82.5\scriptsize{±1.8} &
  96.1\scriptsize{±0.5} &
  41.0\scriptsize{±3.9} &
  82.6\scriptsize{±2.3} \\
 &
  WinCLIP+ &
  94.4\scriptsize{±1.3} &
  {\ul 94.4\scriptsize{±0.8}} &
  {\ul 96.0\scriptsize{±0.3}} &
  {\ul 58.4\scriptsize{±1.7}} &
  {\ul 88.4\scriptsize{±0.9}} &
  84.6\scriptsize{±2.4} &
  {\ul 83.0\scriptsize{±1.4}} &
  96.8\scriptsize{±0.3} &
  {\ul 43.5\scriptsize{±3.3}} &
  {\ul 86.2\scriptsize{±1.4}} \\
 &
  AnomalyGPT &
  95.5\scriptsize{±0.8} &
  - &
  95.6\scriptsize{±0.2} &
  - &
  - &
  {\ul 88.6\scriptsize{±0.7}} &
  - &
  96.4\scriptsize{±0.1} &
  - &
  - \\
 &
  PromptAD &
  {\ul 95.7\scriptsize{±1.5}} &
  - &
  \textbf{96.2\scriptsize{±0.3}} &
  - &
  - &
  88.3\scriptsize{±2.0} &
  - &
  {\ul 97.1\scriptsize{±0.3}} &
  - &
  - \\
 &
  \cellcolor{blue!10}\textbf{Ours} &
  \cellcolor{blue!10}\textbf{95.8\scriptsize{±0.2}} &
  \cellcolor{blue!10}\textbf{95.2\scriptsize{±0.2}} &
  \cellcolor{blue!10}{\ul 96.0\scriptsize{±0.2}} &
  \cellcolor{blue!10}\textbf{58.8\scriptsize{±0.2}} &
  \cellcolor{blue!10}\textbf{92.6\scriptsize{±0.1}} &
  \cellcolor{blue!10}\textbf{91.3\scriptsize{±0.4}} &
  \cellcolor{blue!10}\textbf{87.2\scriptsize{±0.6}} &
  \cellcolor{blue!10}\textbf{97.9\scriptsize{±0.4}} &
  \cellcolor{blue!10}\textbf{44.9\scriptsize{±0.3}} &
  \cellcolor{blue!10}\textbf{92.7\scriptsize{±0.1}} \\ \midrule
\multirow{6}{*}{4-shot} &
  PaDiM &
  80.4\scriptsize{±2.5} &
  90.2\scriptsize{±1.2} &
  92.6\scriptsize{±0.7} &
  46.1\scriptsize{±1.8} &
  81.3\scriptsize{±1.9} &
  72.8\scriptsize{±2.9} &
  78.0\scriptsize{±1.2} &
  93.2\scriptsize{±0.5} &
  24.6\scriptsize{±1.8} &
  72.6\scriptsize{±1.9} \\
 &
  PatchCore &
  88.8\scriptsize{±2.6} &
  92.6\scriptsize{±1.6} &
  94.3\scriptsize{±0.5} &
  55.0\scriptsize{±1.9} &
  84.3\scriptsize{±1.4} &
  85.3\scriptsize{±2.1} &
  {\ul 84.3\scriptsize{±1.3}} &
  96.8\scriptsize{±0.3} &
  43.9\scriptsize{±3.1} &
  84.9\scriptsize{±1.4} \\
 &
  WinCLIP+ &
  95.2\scriptsize{±1.3} &
  {\ul 94.7\scriptsize{±0.8}} &
  {\ul 96.2\scriptsize{±0.3}} &
  {\ul 59.5\scriptsize{±1.8}} &
  {\ul 89.0\scriptsize{±0.8}} &
  87.3\scriptsize{±1.8} &
  84.2\scriptsize{±1.6} &
  97.2\scriptsize{±0.2} &
  {\ul 47.0\scriptsize{±3.0}} &
  {\ul 87.6\scriptsize{±0.9}} \\
 &
  AnomalyGPT &
  96.3\scriptsize{±0.3} &
  - &
  {\ul 96.2\scriptsize{±0.1}} &
  - &
  - &
  {\ul 90.6\scriptsize{±0.7}} &
  - &
  96.7\scriptsize{±0.1} &
  - &
  - \\
 &
  PromptAD &
  \textbf{96.6\scriptsize{±0.9}} &
  - &
  \textbf{96.5\scriptsize{±0.2}} &
  - &
  - &
  89.1\scriptsize{±1.7} &
  - &
  {\ul 97.4\scriptsize{±0.4}} &
  - &
  - \\
 &
  \cellcolor{blue!10}\textbf{Ours} &
  \cellcolor{blue!10}{\ul 96.4\scriptsize{±0.1}} &
  \cellcolor{blue!10}\textbf{95.1\scriptsize{±0.1}} &
  \cellcolor{blue!10}{\ul 96.2\scriptsize{±0.1}} &
  \cellcolor{blue!10}\textbf{59.8\scriptsize{±0.1}} &
  \cellcolor{blue!10}\textbf{93.0\scriptsize{±0.0}} &
  \cellcolor{blue!10}\textbf{91.7\scriptsize{±1.0}} &
  \cellcolor{blue!10}\textbf{87.1\scriptsize{±0.1}} &
  \cellcolor{blue!10}\textbf{97.8\scriptsize{±0.4}} &
  \cellcolor{blue!10}\textbf{47.9\scriptsize{±0.2}} &
  \cellcolor{blue!10}\textbf{93.4\scriptsize{±0.1}} \\ \bottomrule[1.5pt]
\end{tabular}
}
\end{table*}
\begin{table*}[t]
\centering
\caption{\textbf{Comparison of full-shot anomaly detection on MVTec AD and VisA.} The best results are in \textbf{bold}, and the second-best results are {\ul underlined}.}
\label{tab:full_shot}
\renewcommand{\arraystretch}{1.0}
\resizebox{0.8\textwidth}{!}{%
\begin{tabular}{@{}c|ccccc|ccccc@{}}
\toprule[1.5pt]
\multirow{2}{*}{Methods} & \multicolumn{5}{c}{MVTec AD}                                               & \multicolumn{5}{c}{VisA}                           \\ \cmidrule(l){2-11} 
                         & I-AUC   & I-F1    & P-AUC     & P-F1    & PRO           & I-AUC & I-F1 & P-AUC & P-F1 & PRO  \\ \midrule
UniAD                    & 96.5        & {\ul 98.8}  & 96.8          & 43.4        & {\ul 90.7}    & {\ul 88.8}      & {\ul 90.8}     & {\ul 98.3}      & 33.7     & {\ul 85.5} \\
SimpleNet                & 95.3        & 98.4        & {\ul 96.9}    & 45.9        & 86.5          & 87.2      & 87.0       & 96.8      & {\ul 34.7}     & 81.4 \\
DiAD                     & 97.2        & \textbf{99.0} & 96.8          & {\ul 52.6}  & {\ul 90.7}    & 86.8      & 88.3     & 96.0        & 26.1     & 75.2 \\
AnomalyGPT               & {\ul 97.4}  & -           & 93.1          & -           & -             & -         & -        & -         & -        & -    \\
\cellcolor{blue!10} \textbf{Ours}            & \cellcolor{blue!10}\textbf{98.4} & \cellcolor{blue!10}96.9        \cellcolor{blue!10}& \cellcolor{blue!10}\textbf{97.4} & \cellcolor{blue!10}\textbf{65.1} & \cellcolor{blue!10}\textbf{94.7} & \cellcolor{blue!10}\textbf{95.8}         & \cellcolor{blue!10}\textbf{91.9}        & \cellcolor{blue!10}\textbf{98.7}         & \cellcolor{blue!10}\textbf{58.7}        & \cellcolor{blue!10}\textbf{97.7}    \\ \bottomrule[1.5pt]
\end{tabular}
}
\end{table*}

\begin{table}[t]
\centering
\caption{\textbf{Per-category anomaly detection performance on MVTec AD in the 1-shot setup.} We report the mean and standard deviation over 5 random seeds for each measurement.}
\label{tab:mvtec_1shot}
\renewcommand{\arraystretch}{0.8}
\resizebox{\linewidth}{!}{%
\begin{tabular}{cccccc}
\toprule[1.5pt]
Category   & I-AUC                  & I-F1                   & P-AUC                 & P-F1                  & PRO                   \\ \midrule
bottle     & 98.9\scriptsize{±0.1}  & 97.9\scriptsize{±0.5}  & 96.3\scriptsize{±0.0} & 71.0\scriptsize{±0.2} & 92.7\scriptsize{±0.0} \\
cable      & 89.1\scriptsize{±7.6}  & 87.8\scriptsize{±4.5}  & 91.7\scriptsize{±0.0} & 36.4\scriptsize{±2.2} & 82.8\scriptsize{±0.2} \\
capsule    & 93.8\scriptsize{±1.6}  & 94.4\scriptsize{±0.5}  & 97.3\scriptsize{±0.0} & 44.8\scriptsize{±0.0} & 96.2\scriptsize{±0.0} \\
carpet     & 100.0\scriptsize{±0.0} & 100.0\scriptsize{±0.0} & 99.0\scriptsize{±0.0} & 75.1\scriptsize{±0.1} & 97.5\scriptsize{±0.0} \\
grid       & 97.3\scriptsize{±1.3}  & 96.1\scriptsize{±2.2}  & 97.2\scriptsize{±0.0} & 49.8\scriptsize{±0.1} & 92.0\scriptsize{±0.1} \\
hazelnut   & 99.9\scriptsize{±0.0}  & 99.3\scriptsize{±0.2}  & 98.4\scriptsize{±0.0} & 63.5\scriptsize{±0.1} & 97.3\scriptsize{±0.0} \\
leather    & 100.0\scriptsize{±0.0} & 100.0\scriptsize{±0.0} & 99.7\scriptsize{±0.0} & 62.2\scriptsize{±0.2} & 99.5\scriptsize{±0.0} \\
metal\_nut & 95.6\scriptsize{±9.3}  & 95.3\scriptsize{±4.3}  & 91.6\scriptsize{±0.1} & 58.8\scriptsize{±0.5} & 90.7\scriptsize{±0.0} \\
pill       & 93.7\scriptsize{±0.4}  & 96.0\scriptsize{±0.1}  & 94.3\scriptsize{±0.0} & 56.6\scriptsize{±0.2} & 97.0\scriptsize{±0.0} \\
screw      & 74.9\scriptsize{±3.1}  & 87.2\scriptsize{±1.3}  & 97.8\scriptsize{±0.0} & 42.5\scriptsize{±0.3} & 92.3\scriptsize{±0.0} \\
tile       & 99.6\scriptsize{±0.0}  & 98.8\scriptsize{±0.2}  & 95.6\scriptsize{±0.0} & 73.2\scriptsize{±0.0} & 93.3\scriptsize{±0.0} \\
toothbrush & 94.1\scriptsize{±2.7}  & 93.1\scriptsize{±5.6}  & 98.5\scriptsize{±0.0} & 56.6\scriptsize{±1.3} & 94.7\scriptsize{±0.0} \\
transistor & 91.4\scriptsize{±2.1}  & 80.7\scriptsize{±4.4}  & 79.6\scriptsize{±0.0} & 37.1\scriptsize{±0.2} & 64.0\scriptsize{±0.2} \\
wood       & 99.6\scriptsize{±0.0}  & 98.5\scriptsize{±0.1}  & 96.6\scriptsize{±0.0} & 70.0\scriptsize{±0.0} & 96.8\scriptsize{±0.0} \\
zipper     & 96.4\scriptsize{±0.8}  & 96.1\scriptsize{±0.4}  & 97.4\scriptsize{±0.0} & 61.4\scriptsize{±0.4} & 91.4\scriptsize{±0.0} \\ \midrule
\rowcolor{blue!10}Average    & 94.9\scriptsize{±0.4}  & 94.7\scriptsize{±0.4}  & 95.4\scriptsize{±0.2} & 57.3\scriptsize{±0.0} & 91.9\scriptsize{±0.0} \\ \bottomrule[1.5pt]
\end{tabular}
}
\end{table}


\begin{table}[t]
\centering
\caption{\textbf{Per-category anomaly detection performance on MVTec AD in the 2-shot setup.} We report the mean and standard deviation over 5 random seeds for each measurement.}
\vspace{-0.5cm}
\label{tab:mvtec_2shot}
\renewcommand{\arraystretch}{0.8}
\resizebox{\linewidth}{!}{%
\begin{tabular}{cccccc}
\toprule[1.5pt]
Category   & I-AUC                  & I-F1                   & P-AUC                 & P-F1                  & PRO                   \\ \midrule
bottle     & 99.5\scriptsize{±0.1}  & 98.6\scriptsize{±0.1}  & 93.2\scriptsize{±0.1} & 62.3\scriptsize{±1.8} & 88.7\scriptsize{±0.4} \\
cable      & 89.3\scriptsize{±1.4}  & 86.5\scriptsize{±2.0}  & 94.9\scriptsize{±0.0} & 41.1\scriptsize{±0.8} & 86.0\scriptsize{±0.3} \\
capsule    & 95.1\scriptsize{±1.1}  & 94.6\scriptsize{±0.9}  & 96.6\scriptsize{±0.2} & 44.9\scriptsize{±0.0} & 96.1\scriptsize{±0.1} \\
carpet     & 100.0\scriptsize{±0.0} & 100.0\scriptsize{±0.0} & 99.3\scriptsize{±0.0} & 76.5\scriptsize{±0.1} & 98.1\scriptsize{±0.0} \\
grid       & 95.4\scriptsize{±1.2}  & 93.5\scriptsize{±1.9}  & 98.2\scriptsize{±0.0} & 52.2\scriptsize{±0.2} & 93.9\scriptsize{±0.0} \\
hazelnut   & 99.8\scriptsize{±0.0}  & 98.9\scriptsize{±0.1}  & 98.0\scriptsize{±0.0} & 57.4\scriptsize{±0.3} & 96.9\scriptsize{±0.0} \\
leather    & 100.0\scriptsize{±0.0} & 100.0\scriptsize{±0.0} & 99.7\scriptsize{±0.0} & 66.1\scriptsize{±0.3} & 99.5\scriptsize{±0.0} \\
metal\_nut & 99.8\scriptsize{±0.0}  & 99.1\scriptsize{±0.4}  & 94.6\scriptsize{±0.1} & 66.7\scriptsize{±0.5} & 93.2\scriptsize{±0.1} \\
pill       & 96.4\scriptsize{±0.8}  & 96.8\scriptsize{±0.3}  & 94.8\scriptsize{±0.0} & 59.0\scriptsize{±0.3} & 96.9\scriptsize{±0.0} \\
screw      & 78.8\scriptsize{±3.2}  & 88.4\scriptsize{±0.1}  & 98.2\scriptsize{±0.0} & 46.6\scriptsize{±0.2} & 93.5\scriptsize{±0.0} \\
tile       & 100.0\scriptsize{±0.0} & 99.9\scriptsize{±0.1}  & 97.1\scriptsize{±0.0} & 74.5\scriptsize{±0.0} & 94.8\scriptsize{±0.0} \\
toothbrush & 93.4\scriptsize{±0.9}  & 93.1\scriptsize{±0.1}  & 98.7\scriptsize{±0.0} & 59.5\scriptsize{±1.0} & 94.3\scriptsize{±0.0} \\
transistor & 88.9\scriptsize{±2.0}  & 82.5\scriptsize{±4.1}  & 85.1\scriptsize{±0.2} & 41.2\scriptsize{±0.0} & 67.6\scriptsize{±0.7} \\
wood       & 99.4\scriptsize{±0.0}  & 97.7\scriptsize{±0.1}  & 96.9\scriptsize{±0.0} & 70.9\scriptsize{±0.1} & 97.0\scriptsize{±0.0} \\
zipper     & 99.4\scriptsize{±0.0}  & 98.3\scriptsize{±0.4}  & 97.7\scriptsize{±0.0} & 63.6\scriptsize{±1.5} & 92.8\scriptsize{±0.4} \\ \midrule
\rowcolor{blue!10}Average    & 95.8\scriptsize{±0.1}  & 95.2\scriptsize{±0.2}  & 96.0\scriptsize{±0.2} & 58.8\scriptsize{±0.2} & 92.6\scriptsize{±0.1} \\ \bottomrule[1.5pt]
\end{tabular}
}
\end{table}

\begin{table}[t]
\centering
\caption{\textbf{Per-category anomaly detection performance on MVTec AD in the 4-shot setup.} We report the mean and standard deviation over 5 random seeds for each measurement.}
\label{tab:mvtec_4shot}
\renewcommand{\arraystretch}{0.9}
\resizebox{\linewidth}{!}{%
\begin{tabular}{cccccc}
\toprule[1.5pt]
Category   & I-AUC                  & I-F1                   & P-AUC                 & P-F1                  & PRO                   \\ \midrule
bottle     & 99.2\scriptsize{±0.0}  & 98.0\scriptsize{±0.5}  & 96.9\scriptsize{±0.0} & 73.0\scriptsize{±0.1} & 93.7\scriptsize{±0.0} \\
cable      & 90.6\scriptsize{±0.8}  & 87.1\scriptsize{±3.1}  & 93.7\scriptsize{±0.0} & 43.4\scriptsize{±0.9} & 86.6\scriptsize{±0.1} \\
capsule    & 96.4\scriptsize{±0.4}  & 95.8\scriptsize{±0.2}  & 98.1\scriptsize{±0.0} & 46.9\scriptsize{±0.1} & 97.3\scriptsize{±0.0} \\
carpet     & 100.0\scriptsize{±0.0} & 100.0\scriptsize{±0.0} & 99.2\scriptsize{±0.0} & 75.7\scriptsize{±0.0} & 97.8\scriptsize{±0.0} \\
grid       & 98.4\scriptsize{±0.8}  & 96.9\scriptsize{±2.6}  & 98.1\scriptsize{±0.0} & 52.4\scriptsize{±0.0} & 93.6\scriptsize{±0.0} \\
hazelnut   & 98.9\scriptsize{±0.2}  & 97.6\scriptsize{±0.4}  & 98.6\scriptsize{±0.0} & 63.1\scriptsize{±0.3} & 97.3\scriptsize{±0.0} \\
leather    & 100.0\scriptsize{±0.0} & 100.0\scriptsize{±0.0} & 99.7\scriptsize{±0.0} & 63.2\scriptsize{±0.2} & 99.4\scriptsize{±0.0} \\
metal\_nut & 99.3\scriptsize{±0.3}  & 98.4\scriptsize{±0.4}  & 93.2\scriptsize{±0.0} & 64.7\scriptsize{±0.1} & 92.8\scriptsize{±0.0} \\
pill       & 97.3\scriptsize{±0.1}  & 97.4\scriptsize{±0.1}  & 94.9\scriptsize{±0.0} & 59.4\scriptsize{±0.1} & 97.3\scriptsize{±0.0} \\
screw      & 85.5\scriptsize{±0.1}  & 88.6\scriptsize{±0.1}  & 98.3\scriptsize{±0.0} & 49.0\scriptsize{±0.3} & 93.6\scriptsize{±0.0} \\
tile       & 99.8\scriptsize{±0.0}  & 98.7\scriptsize{±0.1}  & 95.9\scriptsize{±0.0} & 73.3\scriptsize{±0.0} & 93.6\scriptsize{±0.0} \\
toothbrush & 94.3\scriptsize{±1.3}  & 94.1\scriptsize{±0.8}  & 98.6\scriptsize{±0.0} & 60.2\scriptsize{±0.9} & 94.2\scriptsize{±0.1} \\
transistor & 87.5\scriptsize{±1.9}  & 75.9\scriptsize{±4.1}  & 82.8\scriptsize{±0.0} & 39.5\scriptsize{±0.2} & 67.4\scriptsize{±0.1} \\
wood       & 99.5\scriptsize{±0.0}  & 97.9\scriptsize{±0.2}  & 96.5\scriptsize{±0.0} & 69.6\scriptsize{±0.0} & 96.7\scriptsize{±0.0} \\
zipper     & 98.8\scriptsize{±0.1}  & 98.5\scriptsize{±0.2}  & 97.9\scriptsize{±0.0} & 64.5\scriptsize{±0.1} & 93.2\scriptsize{±0.0} \\ \midrule
\rowcolor{blue!10}Average    & 96.4\scriptsize{±0.1}  & 95.1\scriptsize{±0.1}  & 96.2\scriptsize{±0.1} & 59.8\scriptsize{±0.1} & 93.0\scriptsize{±0.0} \\ \bottomrule[1.5pt]
\end{tabular}
}
\end{table}

\begin{table}[t]
\centering
\caption{\textbf{Per-category anomaly detection performance on VisA in the 1-shot setup.} We report the mean and standard deviation over 5 random seeds for each measurement.}
\label{tab:visa_1shot}
\renewcommand{\arraystretch}{0.9}
\resizebox{\linewidth}{!}{%
\begin{tabular}{cccccc}
\toprule[1.5pt]
Category  & I-AUC                  & I-F1                  & P-AUC                 & P-F1                   & PRO                   \\ \midrule
candle    & 90.8\scriptsize{±0.3}  & 85.3\scriptsize{±0.7} & 98.9\scriptsize{±0.0} & 37.6\scriptsize{±0.1}  & 98.0\scriptsize{±0.0} \\
capsules  & 91.1\scriptsize{±2.6}  & 88.5\scriptsize{±0.4} & 98.2\scriptsize{±0.0} & 47.9\scriptsize{±2.0}  & 94.2\scriptsize{±0.1} \\
cashew    & 88.9\scriptsize{±16.6} & 88.9\scriptsize{±5.1} & 96.6\scriptsize{±0.1} & 59.3\scriptsize{±0.9}  & 95.6\scriptsize{±0.0} \\
chewinggum  & 97.4\scriptsize{±0.1} & 95.2\scriptsize{±0.3} & 99.6\scriptsize{±0.0} & 77.1\scriptsize{±0.2} & 92.6\scriptsize{±0.1} \\
fryum     & 96.2\scriptsize{±0.9}  & 94.7\scriptsize{±0.8} & 95.4\scriptsize{±0.0} & 40.8\scriptsize{±0.1}  & 92.2\scriptsize{±0.0} \\
macaroni1 & 86.6\scriptsize{±5.2}  & 79.6\scriptsize{±5.9} & 99.7\scriptsize{±0.0} & 30.1\scriptsize{±0.8}  & 96.2\scriptsize{±0.1} \\
macaroni2 & 79.2\scriptsize{±2.1}  & 73.0\scriptsize{±1.2} & 98.4\scriptsize{±0.0} & 28.3\scriptsize{±1.0}  & 90.5\scriptsize{±0.7} \\
pcb1      & 90.8\scriptsize{±2.0}  & 86.1\scriptsize{±1.5} & 98.5\scriptsize{±0.0} & 49.5\scriptsize{±4.6}  & 93.6\scriptsize{±0.0} \\
pcb2      & 84.3\scriptsize{±4.1}  & 78.1\scriptsize{±4.7} & 96.2\scriptsize{±0.0} & 30.8\scriptsize{±0.4}  & 83.7\scriptsize{±0.0} \\
pcb3      & 78.0\scriptsize{±5.2}  & 74.3\scriptsize{±1.7} & 94.9\scriptsize{±0.0} & 32.7\scriptsize{±11.4} & 86.0\scriptsize{±0.0} \\
pcb4      & 96.8\scriptsize{±1.0}  & 92.4\scriptsize{±3.7} & 96.9\scriptsize{±0.0} & 36.1\scriptsize{±0.3}  & 90.2\scriptsize{±0.3} \\
pipe\_fryum & 96.4\scriptsize{±0.6} & 93.1\scriptsize{±2.3} & 98.5\scriptsize{±0.0} & 47.7\scriptsize{±0.1} & 97.6\scriptsize{±0.0} \\ \midrule
\rowcolor{blue!10}Average   & 89.7\scriptsize{±0.8}  & 85.8\scriptsize{±0.5} & 97.7\scriptsize{±0.4} & 43.2\scriptsize{±0.4}  & 92.5\scriptsize{±0.1} \\ \bottomrule[1.5pt]
\end{tabular}
}
\end{table}

\begin{table}[t]
\centering
\caption{\textbf{Per-category anomaly detection performance on VisA in the 2-shot setup.} We report the mean and standard deviation over 5 random seeds for each measurement.}
\label{tab:visa_2shot}
\renewcommand{\arraystretch}{0.9}
\resizebox{\linewidth}{!}{%
\begin{tabular}{cccccc}
\toprule[1.5pt]
Category   & I-AUC                  & I-F1                  & P-AUC                 & P-F1                  & PRO                   \\ \midrule
candle     & 90.5\scriptsize{±0.5}  & 84.4\scriptsize{±0.9} & 98.9\scriptsize{±0.0} & 39.0\scriptsize{±0.1} & 98.0\scriptsize{±0.0} \\
capsules   & 94.5\scriptsize{±1.1}  & 91.0\scriptsize{±1.1} & 98.3\scriptsize{±0.0} & 49.3\scriptsize{±0.5} & 94.5\scriptsize{±0.2} \\
cashew     & 90.7\scriptsize{±0.7}  & 89.5\scriptsize{±0.2} & 97.0\scriptsize{±0.0} & 59.6\scriptsize{±0.1} & 95.9\scriptsize{±0.0} \\
chewinggum & 97.6\scriptsize{±0.2}  & 95.2\scriptsize{±0.5} & 99.5\scriptsize{±0.0} & 76.5\scriptsize{±0.3} & 92.4\scriptsize{±0.0} \\
fryum      & 97.0\scriptsize{±0.1}  & 95.0\scriptsize{±0.2} & 96.1\scriptsize{±0.0} & 44.9\scriptsize{±0.1} & 92.6\scriptsize{±0.1} \\
macaroni1  & 89.6\scriptsize{±0.1}  & 82.4\scriptsize{±0.6} & 99.7\scriptsize{±0.0} & 29.3\scriptsize{±0.3} & 96.3\scriptsize{±0.0} \\
macaroni2  & 77.4\scriptsize{±19.9} & 72.7\scriptsize{±6.9} & 98.4\scriptsize{±0.0} & 28.3\scriptsize{±0.4} & 88.8\scriptsize{±0.0} \\
pcb1       & 92.0\scriptsize{±0.5}  & 87.7\scriptsize{±0.5} & 98.5\scriptsize{±0.0} & 49.3\scriptsize{±3.0} & 93.9\scriptsize{±0.0} \\
pcb2       & 86.7\scriptsize{±2.0}  & 80.2\scriptsize{±2.0} & 96.8\scriptsize{±0.0} & 34.5\scriptsize{±1.9} & 84.5\scriptsize{±0.1} \\
pcb3       & 84.6\scriptsize{±0.4}  & 78.6\scriptsize{±0.2} & 95.4\scriptsize{±0.0} & 40.5\scriptsize{±4.5} & 86.1\scriptsize{±0.1} \\
pcb4       & 97.0\scriptsize{±0.2}  & 93.2\scriptsize{±0.2} & 97.1\scriptsize{±0.0} & 35.9\scriptsize{±0.2} & 91.1\scriptsize{±0.2} \\
pipe\_fryum & 97.9\scriptsize{±0.0} & 95.2\scriptsize{±0.9} & 98.7\scriptsize{±0.0} & 51.2\scriptsize{±0.1} & 97.6\scriptsize{±0.0} \\ \midrule
\rowcolor{blue!10}Average    & 91.3\scriptsize{±0.4}  & 87.2\scriptsize{±0.6} & 97.9\scriptsize{±0.4} & 44.9\scriptsize{±0.3} & 92.7\scriptsize{±0.1} \\ \bottomrule[1.5pt]
\end{tabular}
}
\end{table}

\begin{table}[t]
\centering
\caption{\textbf{Per-category anomaly detection performance on VisA in the 4-shot setup.} We report the mean and standard deviation over 5 random seeds for each measurement.}
\label{tab:visa_4shot}
\renewcommand{\arraystretch}{0.9}
\resizebox{\linewidth}{!}{%
\begin{tabular}{cccccc}
\toprule[1.5pt]
Category   & I-AUC                  & I-F1                   & P-AUC                 & P-F1                  & PRO                   \\ \midrule
candle     & 92.8\scriptsize{±0.3}  & 86.1\scriptsize{±0.9}  & 99.0\scriptsize{±0.0} & 39.8\scriptsize{±0.1} & 98.3\scriptsize{±0.0} \\
capsules   & 94.7\scriptsize{±0.9}  & 91.6\scriptsize{±2.3}  & 98.4\scriptsize{±0.0} & 50.5\scriptsize{±0.3} & 94.6\scriptsize{±0.1} \\
cashew     & 89.9\scriptsize{±13.4} & 89.5\scriptsize{±5.2}  & 97.1\scriptsize{±0.0} & 62.6\scriptsize{±0.3} & 95.8\scriptsize{±0.0} \\
chewinggum & 97.2\scriptsize{±0.3}  & 95.1\scriptsize{±0.1}  & 99.5\scriptsize{±0.0} & 75.8\scriptsize{±0.4} & 92.5\scriptsize{±0.3} \\
fryum      & 96.4\scriptsize{±1.1}  & 93.7\scriptsize{±1.2}  & 96.3\scriptsize{±0.0} & 46.2\scriptsize{±0.1} & 93.1\scriptsize{±0.0} \\
macaroni1  & 89.3\scriptsize{±2.0}  & 82.2\scriptsize{±2.8}  & 99.7\scriptsize{±0.0} & 29.0\scriptsize{±2.2} & 96.8\scriptsize{±0.0} \\
macaroni2  & 79.8\scriptsize{±16.4} & 74.7\scriptsize{±8.8}  & 98.7\scriptsize{±0.0} & 29.0\scriptsize{±0.5} & 90.9\scriptsize{±0.1} \\
pcb1       & 88.3\scriptsize{±11.4} & 81.3\scriptsize{±17.4} & 99.2\scriptsize{±0.0} & 70.1\scriptsize{±1.1} & 93.9\scriptsize{±0.0} \\
pcb2       & 87.8\scriptsize{±1.0}  & 80.5\scriptsize{±1.9}  & 97.1\scriptsize{±0.0} & 35.7\scriptsize{±0.6} & 85.7\scriptsize{±0.1} \\
pcb3       & 89.6\scriptsize{±1.2}  & 82.3\scriptsize{±3.6}  & 96.2\scriptsize{±0.0} & 47.2\scriptsize{±1.3} & 89.0\scriptsize{±0.1} \\
pcb4       & 97.1\scriptsize{±0.6}  & 93.0\scriptsize{±2.3}  & 97.5\scriptsize{±0.0} & 36.9\scriptsize{±0.0} & 92.6\scriptsize{±0.1} \\
pipe\_fryum & 97.3\scriptsize{±1.7} & 94.9\scriptsize{±1.2} & 98.8\scriptsize{±0.0} & 52.3\scriptsize{±0.6} & 97.6\scriptsize{±0.0} \\ \midrule
\rowcolor{blue!10}Average    & 91.7\scriptsize{±1.0}  & 87.1\scriptsize{±0.1}  & 97.8\scriptsize{±0.4} & 47.9\scriptsize{±0.2} & 93.4\scriptsize{±0.1} \\ \bottomrule[1.5pt]
\end{tabular}
}
\end{table}

\begin{table}[t]
\centering
\caption{Per-category anomaly detection performance on \textbf{MVTec AD} in the \textbf{full-shot} setup.}
\label{tab:full_shot_mvtec_per_category}
\setlength{\tabcolsep}{10pt}
\resizebox{\linewidth}{!}{%
\begin{tabular}{cccccc}
\toprule[1.5pt]
Category   & I-AUC & I-F1  & P-AUC & P-F1 & PRO  \\ \midrule
bottle     & 99.7  & 98.4  & 97.5  & 74.9 & 94.7 \\
cable      & 94.9  & 92.6  & 95.9  & 53.0 & 90.6 \\
capsule    & 98.3  & 97.3  & 98.5  & 51.7 & 98.0 \\
carpet     & 100.0 & 99.4  & 99.3  & 75.3 & 98.2 \\
grid       & 99.8  & 99.1  & 99.1  & 52.7 & 96.6 \\
hazelnut   & 99.7  & 98.6  & 99.4  & 76.3 & 98.5 \\
leather    & 100.0 & 100.0 & 99.7  & 60.4 & 99.5 \\
metal\_nut  & 100.0 & 100.0 & 95.8  & 73.9 & 95.2 \\
pill       & 98.5  & 98.3  & 96.3  & 63.6 & 97.9 \\
screw      & 96.6  & 95.1  & 99.0  & 60.6 & 96.0 \\
tile       & 100.0 & 99.4  & 97.2  & 74.6 & 94.8 \\
toothbrush & 96.9  & 95.2  & 99.5  & 71.5 & 97.8 \\
transistor & 94.3  & 84.2  & 88.0  & 45.2 & 70.0 \\
wood       & 99.7  & 98.4  & 97.1  & 70.9 & 97.0 \\
zipper     & 98.0  & 96.7  & 98.8  & 71.5 & 96.3 \\ \midrule
\rowcolor{blue!10}Average    & 98.4  & 96.9  & 97.4  & 65.1 & 94.7 \\ \bottomrule[1.5pt]
\end{tabular}
}
\end{table}

\begin{table}[t]
\centering
\caption{{Per-category anomaly detection performance on \textbf{VisA} in the \textbf{full-shot} setup.}}
\label{tab:full_shot_visa_per_category}
\setlength{\tabcolsep}{10pt}
\resizebox{\linewidth}{!}{%
\begin{tabular}{cccccc}
\toprule[1.5pt]
Category   & I-AUC & I-F1 & P-AUC & P-F1 & PRO  \\ \midrule
candle     & 95.6  & 90.0 & 99.3  & 40.1 & 98.2 \\
capsules   & 96.2  & 93.8 & 99.1  & 60.1 & 96.6 \\
cashew     & 97.4  & 94.5 & 98.2  & 70.4 & 94.6 \\
chewinggum & 98.7  & 97.0 & 99.5  & 75.3 & 91.8 \\
fryum      & 98.4  & 97.5 & 97.4  & 53.6 & 93.9 \\
macaroni1  & 95.3  & 88.5 & 99.9  & 36.4 & 98.5 \\
macaroni2  & 84.7  & 79.3 & 99.6  & 28.9 & 96.7 \\
pcb1       & 95.9  & 91.8 & 98.8  & 41.8 & 95.4 \\
pcb2       & 94.1  & 88.2 & 98.2  & 40.3 & 91.5 \\
pcb3       & 95.9  & 90.4 & 97.5  & 52.9 & 93.4 \\
pcb4       & 99.4  & 96.6 & 98.4  & 46.5 & 94.9 \\
pipe\_fryum & 98.4  & 95.9 & 99.1  & 58.7 & 97.9 \\ \midrule
\rowcolor{blue!10}Average    & 95.8  & 91.9 & 98.7  & 50.4 & 95.3 \\ \bottomrule[1.5pt]
\end{tabular}
}
\end{table}

\begin{table}[ht]
\centering
\caption{t-test on anomaly detection results.}
\vspace{-0.3cm}
\label{tab:ttest}
\setlength\tabcolsep{3.0pt}
\renewcommand{\arraystretch}{1.2}
\resizebox{0.45\textwidth}{!}{%
\begin{tabular}{c|ccccc}
\toprule[1.5pt]
Method & PaDiM &PatchCore & WinCLIP+ &  AnomalyGPT & PromptAD  \\
\hline
p-value & $5e-7$ & $8e-5$ & $1e-7$ & 0.005 & 0.7 \\
  \bottomrule[1.2pt]
\end{tabular}
}
\end{table}



\end{document}